\documentclass[journal]{IEEEtran}
\usepackage{amsmath,amsfonts}
\usepackage{algorithm}
\usepackage[noend]{algpseudocode}
\usepackage{multirow}
\usepackage{array}
\usepackage{booktabs}
\usepackage{threeparttable}
\usepackage[caption=false,font=footnotesize,labelfont=rm,textfont=rm]{subfig}
\usepackage{textcomp}
\usepackage{stfloats}
\usepackage{url}
\usepackage{verbatim}
\usepackage{amssymb} 
\usepackage{indentfirst}
\usepackage{graphicx}
\usepackage{doi}
\usepackage{tabularx}
\usepackage{xcolor}
\graphicspath{{Figs/}} 
\def\BibTeX{{\rm B\kern-.05em{\sc i\kern-.025em b}\kern-.08em
    T\kern-.1667em\lower.7ex\hbox{E}\kern-.125emX}}
\usepackage{balance}
\begin{document}

\title{A Survey of Reinforcement Learning-Based Motion Planning for Autonomous Driving: Lessons Learned from a Driving-Task Perspective}

\author{Zhuoren Li, Guizhe Jin, Ran Yu, Weiqi Zhang, Zhiwen Chen, Nan Li, Lu Xiong,, Ilya Kolmanovsky, Dimitar Filev, Bo Leng, and Jia Hu
\thanks{This work has been submitted to the IEEE for possible publication. Copyright may be transferred without notice, after which this version may no longer be accessible. \textit{(corresponding author: Bo Leng, Jia Hu.)}}
\thanks{Zhuoren Li, Guizhe Jin, Ran Yu, Weiqi Zhang, Zhiwen Chen, Nan Li, Lu Xiong and Bo Leng are with the College of Automotive and Energy Engineering, Tongji University, Shanghai 201804, China. (email: 1911055@tongji.edu.cn)}
\thanks{Jia Hu is with the Key Laboratory of Road and Traffic Engineering of the Ministry of Education, Tongji University, Shanghai 201804, China.}
\thanks{Ilya Kolmanovsky is with the Department of Aerospace Engineering, University of Michigan, Ann Arbor, MI 48109, USA.}
\thanks{Dimitar Filev is with the Hagler Institute for Advanced Study, Texas A\&M University, College Station, TX 77840 USA. }
}

\markboth{Journal of \LaTeX\ Class Files,~Vol.~X, No.~X, August~XXX}%
{Shell \MakeLowercase{\textit{et al.}}: A Sample Article Using IEEEtran.cls for IEEE Journals}

\IEEEpubid{0000--0000/00\$00.00~\copyright~2021 IEEE}

\maketitle

\begin{abstract}
Reinforcement learning (RL), with its ability to explore and optimize policies in complex, dynamic decision-making tasks, has emerged as a promising approach to addressing motion planning (MoP) challenges in autonomous driving (AD). Despite rapid advancements in RL and AD, a systematic description and interpretation of the RL design process tailored to diverse driving tasks remains underdeveloped. This survey provides a comprehensive review of RL-based MoP for AD, focusing on lessons from task-specific perspectives. We first outline the fundamentals of RL methodologies, and then survey their applications in MoP, analyzing scenario-specific features and task requirements to shed light on their influence on RL design choices. Building on this analysis, we summarize key design experiences, extract insights from various driving task applications, and provide guidance for future implementations. Additionally, we examine the frontier challenges in RL-based MoP, review recent efforts to addresse these challenges, and propose strategies for overcoming unresolved issues.
\end{abstract}


\section{Introduction}

\IEEEPARstart{R}{einforcement} learning (RL) is a machine learning paradigm that focuses on solving sequential decision-making and control challenges~\cite{sutton1998reinforcement}. In contrast to supervised learning such as imitation learning (IL)~\cite{ILSurveyNNLS}), where the agent directly learns a policy from expert-labeled data, RL improves a policy through trial-and-error interactions with the environment and maximizes long-term rewards~\cite{I11_RL22}. With RL methods surpassing human world champions in Go~\cite{I12_Nat_go}, Starcraft II~\cite{I14_Nat_SC2}, automobile racing~\cite{I10_Nat_Racing}, and drone racing~\cite{I15_Nat_Drone}, it have motivated increasing interest in applying RL to autonomous driving (AD), particularly to motion planning (MoP)~~\cite{I1,I5}. 

\begin{figure}[!t]
    \centering
    \subfloat[\label{1_a}]{%
        \includegraphics[width=0.26\textwidth]{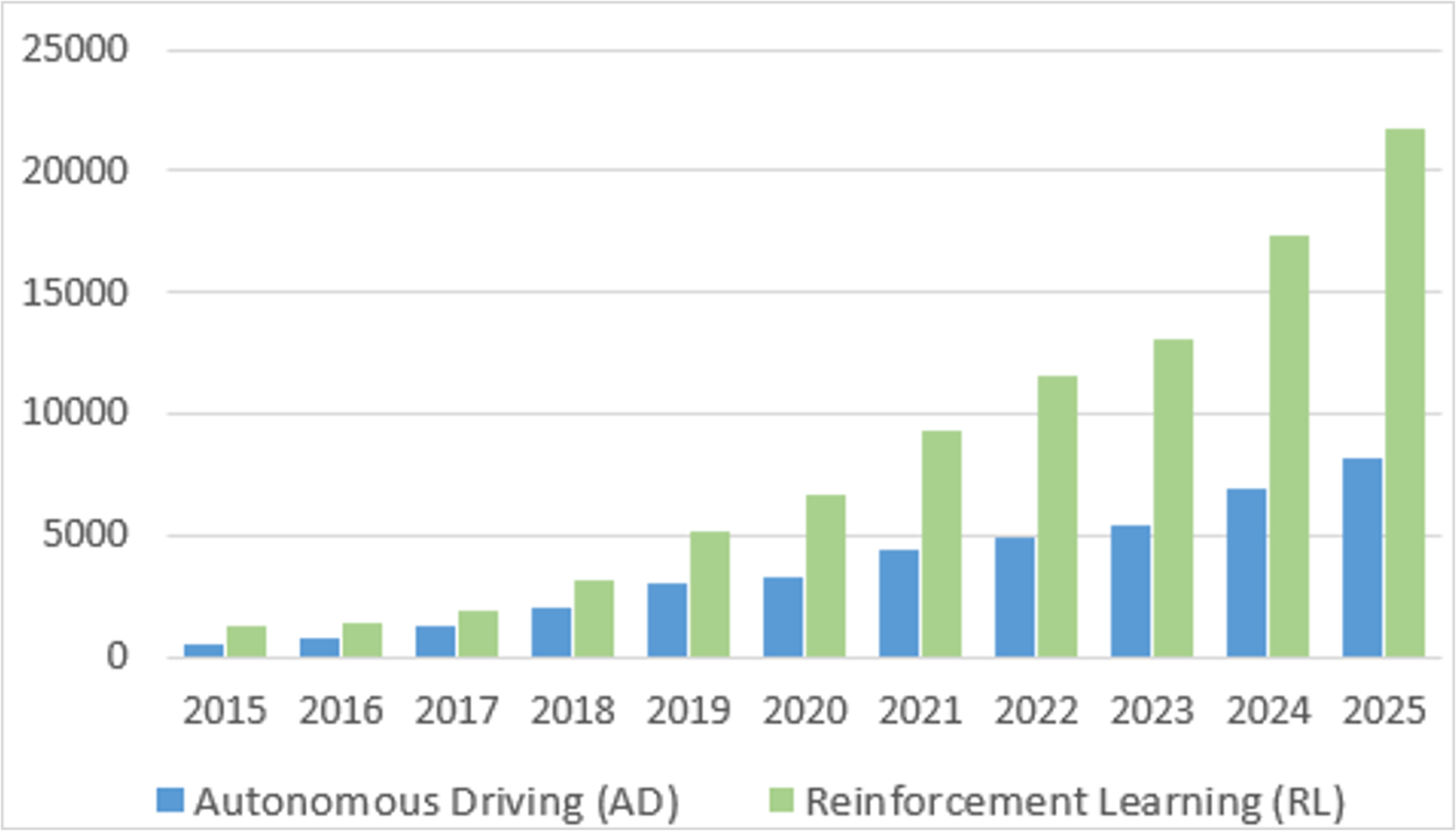}
    }
    \subfloat[\label{1_b}]{%
        \includegraphics[width=0.212\textwidth]{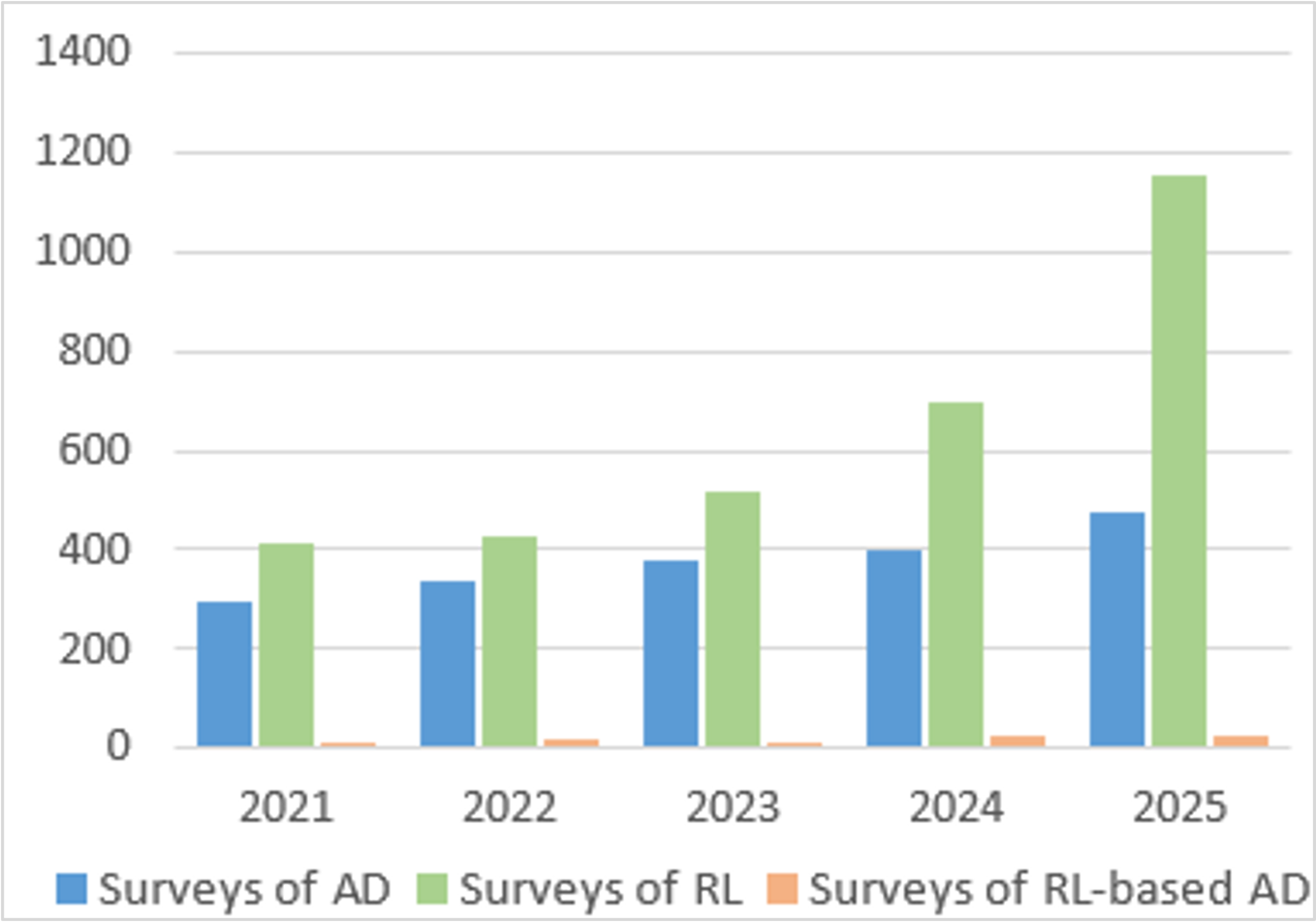}  
    }
    \caption{Search result of Web of Science until 2024: (a) topic search for RL and AD. (b) topic search for surveys for RL, AD, and RL-based AD.}
    \label{fig1}
    \vspace{-1mm}
\end{figure}

According to the statistics from Web of Science (WOS), research on RL and AD has grown rapidly over the past decade, as shown in Fig.~\ref{fig1}. This trend is closely related to the interactive and sequential nature of MoP in AD~\cite{I3_AnnuRev,I2}. Accordingly, RL-based MoP has been explored in a wide range of driving tasks, from lane-level maneuvering to urban navigation and other complex scenarios. 

Most existing surveys on AD either provide broad reviews of the overall system or focus on specific functional modules, such as localization, perception, communication, etc., while MoP has received comparatively less attention~\cite{I8_MileStoneinAD}. Surveys dedicated to RL-based MoP are even more limited. Several studies have provided systematic reviews of representative RL methods and their applications in AD~\cite{I11_RL22,I23gaozhenhai}, offering useful summaries of algorithmic developments and application trends. However, these reviews  are mainly organized from the perspective of RL algorithmic taxonomy and provide limited discussion on the relationship between RL formulation and specific driving tasks. Recent surveys such as~\cite{I7_RL_IL,TRC2024recent} have begun to discuss RL-based MoP from scenario-oriented perspectives, and provide insight into some state-of-the-art RL research from a problem-driven perspective. Nevertheless, their coverage of driving tasks remains limited, with some specialized domains receiving relatively less attention. More importantly, they still lack a systematic analysis of how AD task-specific characteristics and requirements shape key RL design choices. 
\IEEEpubidadjcol
Moreover, the limitations and challenges identified by most surveys, such as driving safety, policy robustness, sample efficiency, and scenario generalization, have been further explored in recent years. Despite the existence of several summaries of advanced theoretical approaches to RL~\cite{I27TPAMI,I28SWRL,I29selfLearn}, to the best of our knowledge, there is no review that comprehensively summarizes the application of these state-of-the-art technologies to the field of MoP for AD. With the rapid development of RL-based AD technologies in both academia and industry, holistic and thorough review of recent investigations is needed.

This article analyzes and summarizes recent advanced work from a comprehensive driving task perspective (although owing to space limitations, we are unable to include some impressive papers in this article). Our study aims to systematically answer the following questions: \textit{How can RL be formulated for different MoP tasks in AD? How do task characteristics influence the RL model design? What recent advances have been made to address the key challenges for RL-based MoP?} The contributions of this article include the following: 

\begin{itemize}
\item We outline the fundamentals of RL methodologies, and then focus on their applications in MoP for AD, where various driving tasks are systematically characterized to shed light on their influence on RL design.
\item We summarize several developments in RL-based MoP for AD, extract insights from various driving task applications, and provide guidance for future implementations.
\item The current challenges in RL applications to MoP for AD are discussed, and beyond pointing out challenges and future directions, a comprehensive review of recent exploratory efforts to address these issues with advanced methods is undertaken.
\end{itemize}

The remainder of this article is organized as follows: Section II briefly introduces the basics of RL and RL-based MoP. Section III reviews research on RL-based MoP from a driving task perspective. Section IV discusses the lessons learned from RL-based MoP design for various driving tasks, and offers experiences and insights. Section V analyzes the current challenges in RL-based MoP and details exploratory efforts to apply advanced RL theories to address them, exploring outlooks and opportunities. Section VI concludes this article.

\section{Basics of RL and RL-based MoP for AD}
\subsection{Basic Theory and Algorithm of Reinforcement Learning}

RL formulates sequential decision-making as an interaction process between an agent and its environment. At each time step, the agent observes the environment state, selects an action, and receives a reward signal that evaluates the consequence of this action. Different from IL, which learns policies from expert-labeled demonstrations, RL improves the policy by maximizing cumulative rewards through environment interactions. A standard RL problem is commonly modeled as a Markov Decision Process (MDP)~\cite{b1MDPNNLS} defined by a tuple $< \mathcal{S},\mathcal{A},\mathcal{R},\mathcal{T},\gamma >$:
\begin{itemize}
\item $\mathcal{S}$ and $\mathcal{A}$ denote the state and action spaces, respectively, i.e. $s_t \in \mathcal{S}$ and  $a_t \in \mathcal{A}$.
\item $\mathcal{T}$ : $\mathcal{S} \times \mathcal{A} \to [0, 1]$, $\mathcal{T}(s_{t+1}, s_t, a_t)$ is the transition function from a current state-action pair $(s_t, a_t)$ to a new state $s_{t+1}$ at the next time step with probability $P(s_{t+1} \mid s_t, a_t)$, which is referred to as the environmental dynamics (system dynamics). 
\item $\mathcal{R}:\mathcal{S}\times \mathcal{A} \times \mathcal{S} \to \mathbb{R}$ is the reward function used to evaluate the agent’s performance.
\item $\gamma \in [0,1] $ denotes the discount factor for the present value of the future reward.
\end{itemize}

When the full environment state is not directly observable, the MDP can be extended to a partially observable Markov decision process (POMDP)~\cite{B2POMDP}. In addition to the MDP components, a POMDP introduces an observation space $\mathcal{O}$ and an observation function $\Omega:\mathcal{S}\times\mathcal{A}\rightarrow \mathcal{O}$, where the agent receives an observation $o_{t+1}$ according to the observation probability $P(o_{t+1}|s_{t+1},a_t)$ rather than directly accessing the underlying state.

The policy $\pi:(a_t | s_t)$, maps the observed state $s_t$ to a probability of an action $a_t$, which represents the driving maneuver in the AD driving task. The set of all possible policies is expressed by $\Pi$. The sequence $\{s_0, a_0, s_1, a_1, \cdots , s_t, a_t,  \cdots \}$ generated by the RL agent with the policy $\pi$ is called trajectory or rollout. The solution objective of the MDP is to find the optimal policy $\pi^{*}$ resulting in the highest expected discounted return  over all possible trajectories, where $h$ is the current timestep and $H$ is the finite horizon (for an infinite horizon $H$ is set to $\infty$). Furthermore, the expectation of return following the policy $\pi$ from a state $s$ is defined as the value-function:
\vspace{-2mm}
\begin{equation}
V_\pi (s)=\mathbb{E}[G_t|s_t = s]=\mathbb{E}[\sum_{t=h}^{h+H}\gamma ^{t-h}\mathcal{R}_{t+1}|s_h=s ]
\label{b1}
\end{equation}

\noindent where $G_t$ means the total return for the current state $s_t$. Similarly, the action-value function, i.e., “Q-value function” is defined as:
\begin{equation}
\begin{aligned}[b]
&Q_{\pi}(s_t,a_t)=\mathbb{E}[G_t|s_t = s,a_t=a]\\&=\mathbb{E}[\mathcal{R}_{t}+\gamma Q_{\pi}(s_{t+1},\pi(a_{t+1})|s_{t+1})]
\label{b2}
\end{aligned}
\end{equation}

RL methods can be broadly classified into model-based and model-free approaches according to whether an explicit transition model is used. Model-based RL relies on known or learned environment dynamics for planning or policy optimization, such as dynamic programming (DP)~\cite{B3DP}. However, accurate transition dynamics are difficult to obtain in AD MoP tasks with complex traffic interactions, which restricts its applicability~\cite{B4MBRL}. Model-free RL instead learns value functions or policies directly from sampled interactions, commonly through Monte Carlo or temporal-difference learning~\cite{B5ebe,B6TD}, without explicitly modeling the transition function. For high-dimensional or continuous state spaces, deep neural networks (DNNs) are widely used to approximate value functions, policies, or both, where $\pi_\theta$ denotes a policy parameterized by network parameters $\theta$.

\begin{figure}[!t]
    \centering
    \subfloat[Categorized by policy generation paradigm.\label{2_a}]{%
        \includegraphics[width=0.45\textwidth]{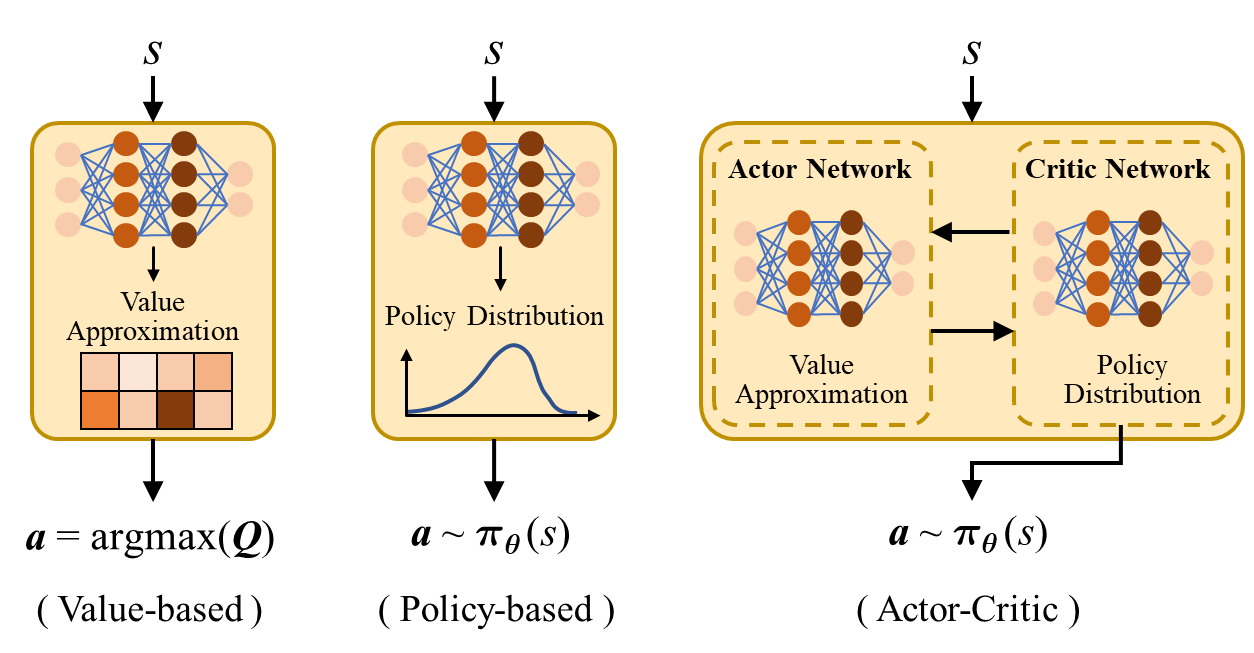}
    }
    \\
    \subfloat[Categorized by agent interaction configuration.\label{2_b}]{%
        \includegraphics[width=0.45\textwidth]{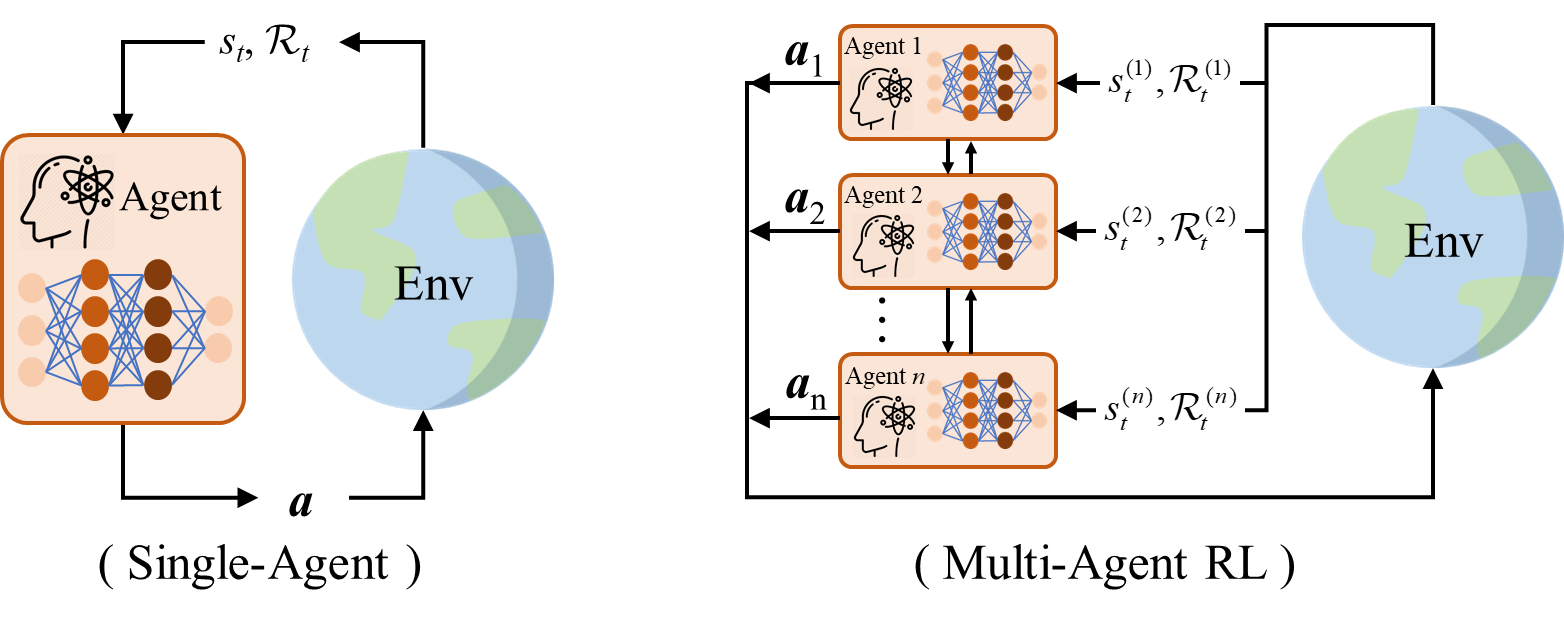}  
    }
    \\
    \subfloat[ Categorized by learning mode.\label{2_c}]{%
        \includegraphics[width=0.48\textwidth]{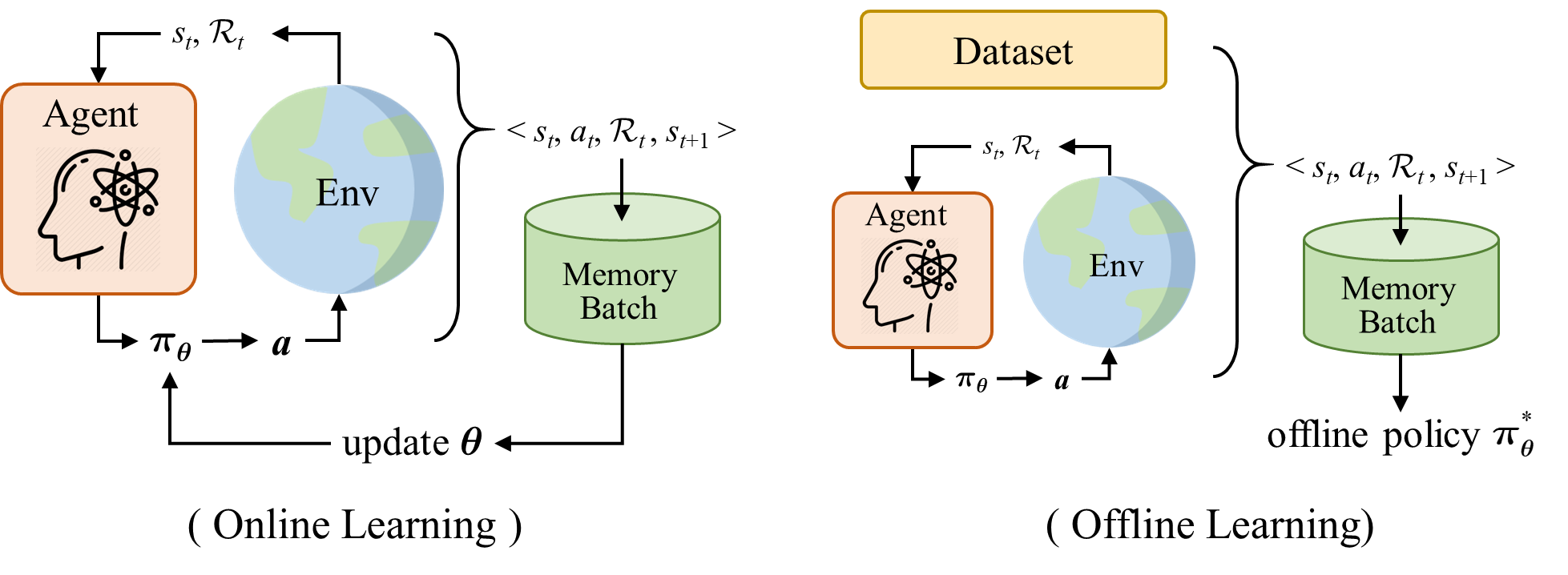}  
    }
    \caption{RL methods with different categorization.}
    \label{f2}
    \vspace{-1mm}
\end{figure}

In this article, RL algorithms relevant to MoP are summarized from three non-mutually-exclusive perspectives: policy generation paradigm, agent interaction configuration, and learning mode, as shown in Fig.~\ref{f2}. Since most RL-based MoP studies in AD adopt model-free deep RL due to the difficulty of obtaining accurate transition dynamics, the following discussion mainly focuses on model-free methods.

\textbf{\textit{1) Policy Generation Paradigm}}

\textit{a) Value-based Methods:} These methods explicitly learn an action-value function and derive the policy by selecting actions with high estimated values. Q-learning is a representative value-based method, whose optimal policy aims to maximize the Q-value:
\vspace{-1mm}
\begin{equation}
\begin{aligned}[b]
\begin{aligned}
&\arg\max_{\pi}Q_{\pi}(s_t,a_t)=\arg\max_{\pi}\mathbb{E}\left [\sum_{t=h}^{h+H}\gamma^t \mathcal{R}(s_t,\pi (a_t|s_t))   \right ]
\end{aligned}
\label{VB1}
\end{aligned}
\end{equation}

The RL agent can update their policies by estimating Q-value as follows: 
\vspace{-1mm}
\begin{equation}
\begin{aligned}[b]
\begin{aligned}
&Q_{\pi}(s_t,a_t)\gets Q_{\pi}(s_t,a_t)\\&+\alpha \left [\mathcal{R}_t+\gamma \max_{a_{t+1}\in\mathcal{A} } Q_{\pi}(s_{t+1},a_{t+1})-Q_{\pi}(s_{t},a_{t}) \right]
\end{aligned}
\label{VB2}
\end{aligned}
\end{equation}

\noindent where $\alpha$ is the learning rate. With the use of DNNs, Q-learning has evolved into far-reaching algorithms represented by Deep Q-Network (DQN)~\cite{I9_Nat_15}, Double DQN (DDQN)~\cite{Ba1ddqn}, dueling DQN~\cite{Ba1dueling}.

The loss function of the Q-network in the DQN can be expressed as:
\begin{equation}
\begin{aligned}[b]
\begin{aligned}
l_{t}^{Q}(\theta)\!=\!\frac{1}{2}\!\left[\mathcal{R}_{t}\!+\!\gamma\!\max _{a_{t+1} \in \mathcal{A}} Q^{\prime}\!\left(s_{t+\!1}, a_{t+1}\! ; \theta^{\prime}\right)\!-\!Q\!\left(s_{t}, a_{t} ; \theta\right)\right]\!^{2}
\end{aligned}
\label{VB3}
\end{aligned}
\end{equation}
where $Q^{\prime}(\cdot;\theta^{\prime})$ s the target Q-network parameterized by $\theta^{\prime}$. Furthermore, DDQN reduces overestimation bias by decoupling action selection and value evaluation, while dueling DQN separately estimates the state value and action advantage. D3QN combines these improvements.

\textit{b) Policy-based Methods:} Unlike value-based methods that derive policies indirectly from value functions, policy-based methods directly optimize a differentiable parameterized policy. They are particularly suitable for continuous control problems with large or continuous action spaces. For a stochastic policy $\pi_\theta$, the optimization objective can be written as:
\begin{equation}
J(\theta)=\mathbb{E}_{\pi_{\theta}}\left[G_{t} \mid s_{t}=s\right].
\label{PB1}
\end{equation}
\indent Policy gradient methods~\cite{B5ebe} use gradient descent to estimate the policy parameters that maximize the expected reward:
\vspace{-1mm}
\begin{equation}
\begin{aligned}
\nabla J(\theta )=&\mathbb{E}\left [ \sum_{t=h}^{h+H}\gamma ^t \mathcal{R}_t (s_t,\pi_\theta (a_t|s_t))\nabla \log\pi_\theta (a_t|s_t))   \right ]\\
&\theta \gets \theta + \nabla J(\theta )
\end{aligned}
\label{PB2}
\end{equation}

The REINFORCE algorithm~\cite{ba2reinforce} uses the Monte Carlo method to estimate $Q_{\pi}(s_t,a_t)$, but the estimation results exhibit large variance. In addition, the advantage function $A_{\pi}(s_t,a_t)=Q_{\pi}(s_t,a_t)-V_{\pi}(s_t,a_t)$~\cite{ba2AF} can be utilized to replace $Q_{\pi}(s_t,a_t)$ to emphasize better actions. Note that policy-based methods can also use a deterministic policy (determining an action based on the state $s$, i.e. $a=\mu_{\theta}(s)$, which can be more efficient), rather than just a stochastic policy (selecting an action from a probability distribution, $a\sim\pi_{\theta}(\cdot|s)$). In this case, the gradient of the objective function can be expressed as:
\begin{equation}
\begin{aligned}
\nabla_{\theta} J(\mu_{\theta})=\mathbb{E}_{\mu_{\theta}}\left [\nabla_{\theta}\mu_{\theta}(s)\nabla_aQ_{\mu_{\theta}}(s_{t+1},a_t|_{a_t=\mu_{\theta}})  \right ]
\end{aligned}
\label{PB3}
\end{equation}

Since deterministic policies do not directly induce stochastic exploration, Deterministic Policy Gradient (DPG)~\cite{ba2DPG} is usually implemented in an off-policy manner with exploratory behavior policies.

\textit{c) Actor-Critic Methods:} Actor-Critic methods are a special type of policy-based method that integrates techniques from value-based methods, where the actor is the policy function $\pi_\theta$ generating actions to obtain the maximum return, and the critic is the value function $V_{\pi_\theta}$ that estimates the actions. This coupled structure integrates the flexibility of policy optimization and the stability of value estimation. The Deep Deterministic Policy Gradient (DDPG)~\cite{DDPG}, Proximal Policy Optimization (PPO)~\cite{PPO}, and Soft Actor-Critic (SAC)~\cite{SAC} algorithms are typical algorithms that utilize the actor-critic framework. Among them, PPO is widely adopted owing to its stable policy updates and implementation simplicity, while SAC achieves high sample efficiency through off-policy learning and entropy regularization. Consequently, both PPO and SAC have become dominant paradigms in recent RL-based MoP studies~\cite{RLSNNLS24}.

\textbf{\textit{2) Agent Interaction Configuration}}

According to the number of agents interacting with the environment, RL methods can be broadly categorized into single-agent RL (SARL) and multi-agent RL (MARL).

In a SARL, a single agent learns a policy based on its own observations and rewards. Most RL-based MoP studies formulate the ego vehicle as the only learning agent, while surrounding vehicles, pedestrians, and traffic signals are treated as part of the environment. This formulation is simple and widely applicable to various driving tasks of single AD. 

MARL allows multiple agents to interact with a shared environment, where each agent makes decisions based on its own observations while being influenced by the joint actions of others. Such interactions are commonly formulated as a Markov Game (MG)~\cite{Bb6Exploration}, which is defined by an extension tuple $< \mathcal{S}, N, {\mathcal{A}^{(i)}}_{i=1\sim N}, {\mathcal{R}^{(i)}}_{i=1\sim N}, \mathcal{T}, \gamma, \Omega, {\mathcal{O}^{(i)}}_{i=1\sim N}>$, where ${\mathcal{A}^{(i)}}_{i=1\sim N}$ is the action sets for N agents, ${\mathcal{R}^{(i)}}_{i=1\sim N}$ is the reward set and $\mathcal{T}: \mathcal{S} \times \mathcal{A}^{(1)} \times \cdots \times \mathcal{A}^{(N)} \to [0,1]$ is the transition function. Each agent receives a local observation $\mathcal{O}^{(i)}$ by $\Omega(\mathcal{S}, i)$ and the relationships between agents can be categorized as cooperative, competitive, or mixed~\cite{Bb7marl}. The training and execution paradigms of MARL can be broadly categorized as centralized or decentralized. Since fully centralized execution is difficult in real-world AD due to the need for real-time communication and shared control~\cite{Bb8marlsv}, most studies adopt centralized training with decentralized execution, where global information is available during training but each agent acts independently, or decentralized training with decentralized execution, where agents learn and act based only on local information. However, MARL still faces scalability challenges because the joint state-action space grows rapidly with the number of agents, reducing training efficiency~\cite{Bb9marl}.

\textbf{\textit{3) Learning Mode}}

Online RL allows the agent to freely interact with the environment and thus collect experience. The RL agent is required to collect sample data (trial-and-error experience) by itself in the training environment and relies on these data to update the policy. This allows the RL agent to discover an unknown optimal policy. However, online RL usually suffers from sample inefficiency in some tasks and places high demands on the fidelity of the training environment. 

Offline RL is a framework dedicated to policy optimization from static, previously collected datasets, and it capitalizes on historical interaction data to derive optimal policy. In contrast to online RL, Offline RL relies solely on a pre-established dataset $\mathcal{D}$, thereby eliminating the need for ongoing exploration while mitigating associated risks. The core objective of offline RL is to minimize the Bellman error:
\vspace{0.5mm}
\begin{equation}
\begin{aligned}[b]
\begin{aligned}
\nabla\!J(\theta)\!&=\!\mathbb{E}_{s_t\!,a_t\!,s_{t+1\!}\sim\!\mathcal{D}}[\mathcal{R}_t\!+\!\gamma\mathbb{E}_{a_{t+1}\sim\pi_{\text{off}}} [(Q^{\pi_\theta}(s_{t+1},a_{t+1})]\\&-Q^{\pi_\theta}(s_{t},a_{t}))^2] 
\end{aligned}
\label{OFFRL1}
\end{aligned}
\end{equation}

Achieving accurate error estimation requires alignment between the evaluation policy and the target policy. However, offline RL inherently aims to discover policies that outperform the original policy, which introduces an unavoidable distributional shift. This shift occurs when the state-action distribution under the learned policy diverges from that under the original policy, leading to inaccuracies in value estimation due to cumulative biases from sampling and function approximation.

To address this distributional shift, Offline RL methods are broadly divided into model-based and model-free methods. Model-based methods leverage learned dynamics models to estimate uncertainty and handle distributional discrepancies. Prominent examples include MORel~\cite{Bc1MORel}, MOPO~\cite{Bc2MOPO}, COMBO~\cite{Bc3COMBO}, etc. Model-free methods are further split into explicit and implicit regularization techniques. Explicit regularization methods, such as (Batch-Constrained Q-learning) BCQ~\cite{Bc3BCQ}, (Bootstrapping Error Accumulation Reduction) BEAR~\cite{Bc4BEAR}, Conservative Q-Learning (CQL)~\cite{Bc5CQL}, etc., impose direct constraints on policy improvement to limit distributional divergence and encourage conservative policy update.

Additionally, the inability to interact with the environment to find more rewarding regions further restricts the performance of offline RL.

\subsection{RL-based Motion Planning for Autonomous Driving} 
\begin{figure}[!t]
    \centering
    \includegraphics[width=0.49\textwidth]{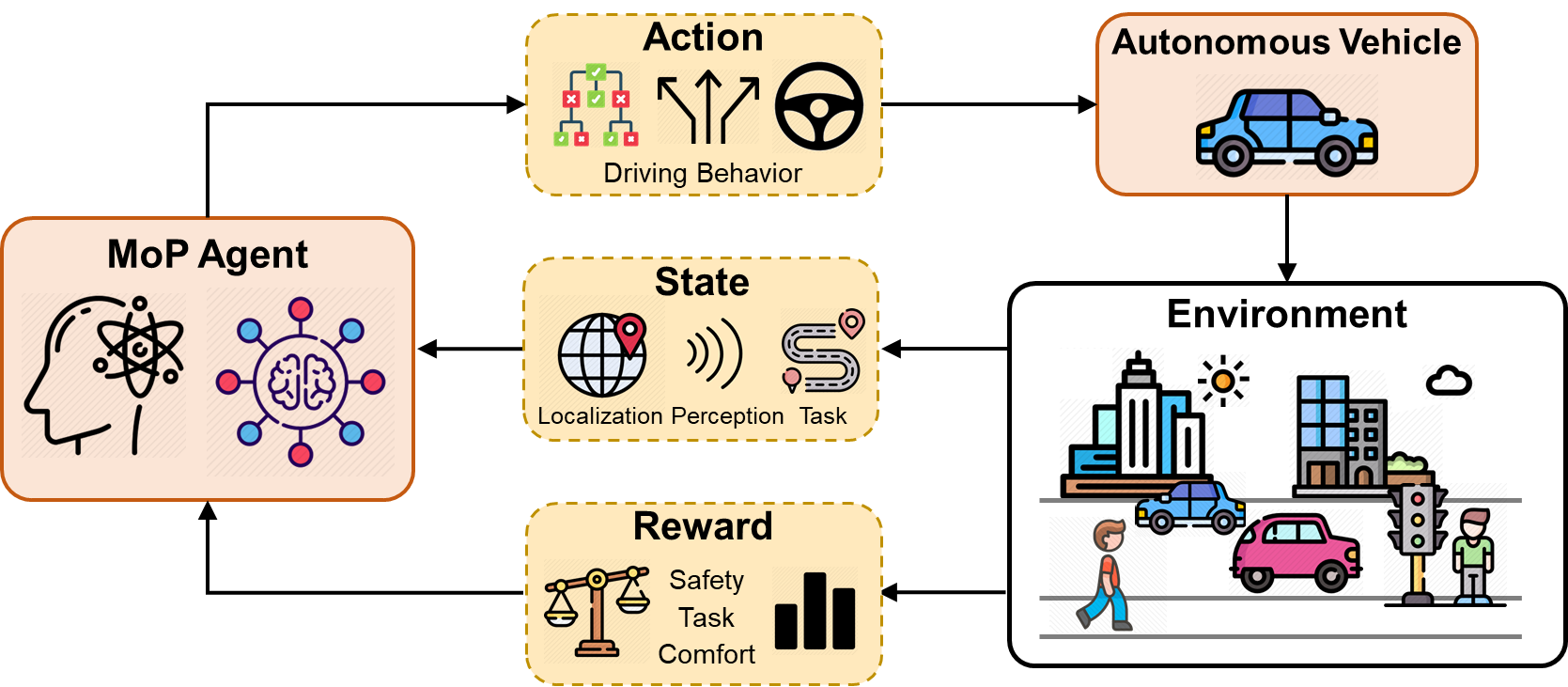}
    \caption{RL algorithm applied to MoP for AD.}
    \label{fig3}
    \vspace{-2mm}
\end{figure}

MoP for AD aims to generate feasible states, trajectories or control sequences, enabling vehicles to safely and efficiently complete driving tasks under dynamic environmental constraints~\cite{I4}. As illustrated in Fig.~\ref{fig3}, RL formulates MoP as a sequential decision-making problem, where the ego vehicle (EV) states, environmental observations, route information, or task objectives constitute the state space, and the learned policy outputs behavioral decisions, trajectory-level commands, or low-level control actions.

Value-based methods are widely used for behavioral planning in MoP~\cite{I6lisrl,vbrlmp4,vbrlmp9}, where discrete driving behavior such as lane keeping, lane changing, yielding, or acceleration selection can be implemented by downstream tracking controllers~\cite{vbrlmp1}. Policy-based methods are more suitable for continuous motion generation such as the steering angle and acceleration~\cite{pbrlmp5,pbrlmp6}. Since the superior performance of the actor-critic methods, they have become widely adopted for learning vehicle control commands directly~\cite{I18sc,pbrlmp4,SACMP1}. Hierarchical RL architectures further decomposes MoP into high-level decision-making and low-level trajectory or control generation. Some works \cite{HA01,LC7_4,In2_1,stateenhanced} train the high-level policy to select discrete semantic decision actions, and then utilize a separate low-level policy to directly control the steering angle and acceleration, achieving more precise and flexible motion control while ensuring clear driving objectives. For highly interactive scenarios, MARL is often used to model multi-vehicle coordination and negotiation, such as cooperative lane changing, merging, and formation control~\cite{CTDE1, CTDE4}.

The key advantage of RL for MoP lies in its ability to optimize long-term closed-loop returns through interaction, rather than merely imitating observed behaviors or following predefined rules. This property enables RL to potentially outperform human drivers by uncovering innovative driving policies that extend beyond traditional rule-based models.

\section{A review from the driving task perspective}{\label{review}}
Most RL-based MoP studies in AD focus on specific driving tasks, ranging from lane keeping and car following to integrated urban navigation. These tasks differ in scenario complexity, interaction patterns, and motion objectives, which directly affects the design of RL models. Therefore, this section first analyzes the scenario characteristics and task requirements of representative driving tasks, and then reviews how RL-based MoP methods design observations, actions, rewards, and learning architectures for these tasks.

\subsection{Lane-Level Maneuvering}
Lane-level maneuvering represents a fundamental class of RL-based MoP tasks on structured roads, including basic longitudinal/lateral tracking and more interactive lane-changing or overtaking behaviors. 

Car following (CF) and lane keeping (LK) are two early and relatively simple RL applications. CF aims to regulate the longitudinal motion of the ego vehicle w.r.t. the front vehicle, typically through continual acceleration~\cite{CF1}, discrete acceleration interval~\cite{vbrlmp9}, or target-speed~\cite{CTDE1} actions, while LK focuses on lateral tracking along the lane centerline or a reference path, where steering-related commands are commonly used~\cite{vbrlmp4,LK1}. Since both tasks involve relatively simple interaction structures and well-defined objectives, compact physical feature states, such as vehicle speed, relative speed, headway distance, steering angle, and preview-path errors, are often sufficient for environment representation and policy learning. Their rewards are usually designed around desired speed and distance, and also include surrogate safety metrics such as Time to Collision (TTC), Time Headway (THW), etc. 

Lane changing (LC) and overtaking involve stronger interaction coupling and higher safety criticality. LC is a common maneuver but also a major source of collision risk~\cite{LCaccident}. Overtaking can be viewed as a sequence of lane-level maneuvers, including moving to the passing lane, passing the preceding vehicle, and returning to the original lane~\cite{LC1}. Therefore, the agent must reason simultaneously about surrounding-vehicle interactions, target-lane availability, and safety-gap acceptance, which substantially increases planning complexity.

Many studies formulate LC and overtaking as high-level behavior planning problems, where DQN and its variants output semantic actions such as LCL, LCR, and LK~\cite{vbrlmp1,LC7_4,LC7_5,LCrealworld}. Among them,~\cite{LCrealworld} presents a real-world application of an RL-based LC policy, while~\cite{RL_GT} further incorporates ac/deceleration into the action space. However, such discrete actions limit maneuver flexibility. Recent studies adopt actor-critic methods such as DDPG, PPO, and SAC to generate steering and acceleration commands~\cite{I18sc,SACMP1,RISKINPUT}, enabling continuous control but also increasing action instability. To balance flexibility and execution stability, trajectory-level or target-level actions have been introduced as intermediate representations. For example, Yu et al.~\cite{LC1} select trajectories from a discrete trajectory set, while Lu et al.~\cite{LC22} output a target point and desired speed for downstream motion optimization. Such intermediate actions provide a compromise among maneuver flexibility, policy interpretability, and control stability.

The increased interaction complexity also motivates richer observation representations. LC and overtaking require simultaneous reasoning about multiple surrounding vehicles and potential future conflicts. Consequently, many studies employ physical feature states of both the EV and surrounding vehicles (SVs), such as the states of the nearest six~\cite{RL_GT} or eight~\cite{I6lisrl} vehicles, or all vehicles within a predefined observation range. To better capture spatial occupancy and interaction context,~\cite{I18sc} further represents the surrounding environment using a discrete state grid. More recently, LiDAR~\cite{LC12_1}, camera~\cite{pbrlmp2}, and other sensor inputs have been introduced to enable End-to-End (E2E) decision making and control. 

Despite the diversity of state and action designs, reward formulations for LC and overtaking relatively consistent since similar task objectives. Safety is the primary concern and is typically represented through collision penalties~\cite{stateenhanced}, relative-distance~\cite{vbrlmp1}, TTC~\cite{CF1}, and other indicators. Efficiency rewards are typically associated with vehicle speed~\cite{LC7_5}, traffic progress, or task completion~\cite{LC20}. Additional reward terms are often introduced to improve comfort by penalizing excessive acceleration and jerk~\cite{LC21}, while traffic-rule rewards encourage compliant behaviors, such as legal overtaking and proper lane usage~\cite{LK1}. Furthermore, some studies design phase-specific rewards according to the overtaking process to explicitly represent the sub-objectives of lane departure, passing, and lane returning~\cite{LC1}, encouraging the agent to follow a desired driving pattern.

\subsection{Conflict-Zone Interaction}

Compared with lane-level maneuvering, conflict-zone interaction tasks are characterized by explicit future path conflicts among surrounding traffic participants. The future motion of the EV and SVs may overlap within a specific conflict region, requiring the EV to make timely operation. Therefore, the core challenge shifts from ego-centered lane-level motion generation to interaction-aware priority negotiation among multiple traffic participants. Typical conflict-zone interaction tasks in MoP include ramp merging, intersection, and roundabout driving.
In ramp merging, the EV needs to adjust its speed before reaching the merge point to find an acceptable gap in the target lane~\cite{RM1,RM2}. Intersections involve more diverse conflict patterns, such as turning versus going straight, unsignalized priority negotiation, and complex traffic-rule constraints, making them among the most challenging structured-road tasks for MoP~\cite{TRC2024recent}. Roundabouts further combine merging, intersection-like negotiation and lane changing, creating a multi-task conflict-zone scenario that poses greater challenges for task-specific context understanding~\cite{I18sc}.

Early RL methods usually adopt longitudinal control or decision-level actions to reduce the complexity of interaction. For example, the agent can learn acceleration/deceleration behavior~\cite{RM3} or speed control commands~\cite{giveaway}, so that the EV reaches the merge region at an appropriate time and finds an acceptable gap. Some merge settings further allow the EV to complete the merge anywhere between a start and an end merge point~\cite{RM5}. For intersections, similar longitudinal-control formulations have been used to adjust acceleration/deceleration~\cite{In1}, while other studies abstract the action space into priority-related decisions, such as wait, pass, yield, take up, and give up~\cite{In2_1,In2_2,rightway}. To improve maneuver flexibility, later studies allow the agent to directly output steering angle and acceleration for ramp merging~\cite{pbrlmp4} or intersection driving~\cite{In3_1,In3_2}. Furthermore, in~\cite{I18sc}, the action space could combine macro-scale lane-changing decisions with mesoscopic acceleration and action-duration commands. Since these scenarios involve stronger coupling among traffic participants, several studies explicitly formulate them as multi-agent interaction problems. Game-theoretic models are employed to describe interaction behaviors~\cite{RM9}, while more recent studies adopt MARL frameworks, where each vehicle is modeled as an independent agent with its own policy for cooperative driving. To further reduce decision complexity, some studies decompose intersection negotiation into multiple sub-tasks~\cite{In5} or employ state machines to manage different interaction modes~\cite{In6}.

The observation design in conflict-zone tasks also becomes more topology- and context-aware. Compared with lane-level maneuvering, the agent must not only observe surrounding vehicles but also understand conflict-region geometry, priority relations, traffic rules, and task progress. Road geometry is usually introduced as additional observation information since merge points, stop lines, and exit positions directly affect the timing and feasibility of entering/existing the conflict region~\cite{RM1,RM2}. Increasing scenario complexity motivates richer observation inputs, including LiDAR point clouds for extracting features of vehicles, bicycles, and pedestrians~\cite{In3_2}, BEV representations from multi-view camera images~\cite{SACMP1}, and roadside perception information to compensate for occlusion~\cite{In4_3,In9_2}. Specially, Zhang et al.~\cite{I18sc} explicitly divide the state input into an environmental representation and a task representation, where the former encodes surrounding-vehicle features and the latter includes relative lane and exit distances. This design reflects the need to represent both traffic interactions and task-specific context in multi-task conflict-zone scenarios.

Reward design in conflict-zone tasks is closely tied to successful task completion, because the agent must accomplish explicit interaction objectives such as merging into the target lane, passing through the intersection, or exiting the roundabout correctly. Therefore, task-completion rewards are commonly introduced to guide the policy toward the desired conflict-resolution outcome. For ramp merging, additional rewards are often assigned for successfully reaching the target lane~\cite{RM7_2}, while driving distance or time is used to evaluate merge efficiency~\cite{RM3,RM8_2}. For intersection driving, reward functions commonly reflect success rate, passage time, collision avoidance, and tracking error~\cite{In3_1,In3_2}, and some studies further penalize traffic-rule violations such as crossing solid lines or running red lights~\cite{In4_3,fearsrl}.

\subsection{Task-Specific Driving Domains}

Some driving tasks are primarily driven by task-specific constraints rather than regular traffic-flow interactions. Parking, racing, and off-road driving belong to this category, but their dominant requirements differ substantially. Parking emphasizes low-speed geometric precision in constrained spaces, racing focuses on high-speed performance and vehicle-dynamics limits, while off-road driving requires terrain-aware control and adaptation without explicit road bound. Therefore, RL design in these domains is usually shaped by the corresponding task objective, such as accurate terminal pose reaching, lap-time, and terrain traversability.

Parking scenarios usually involve partially structured urban environments with perpendicular, parallel, or diagonal parking slots. The main objective is to guide the vehicle to a target position and heading within a constrained space, while avoiding obstacles and reducing unnecessary gear shifts in narrow or unconventional parking spaces~\cite{Park01}. Accordingly, most RL-based parking studies directly control vehicle motion and use the position, velocity, and heading angle of the EV as essential observations. Rewards are typically designed around collision avoidance, target-position and heading accuracy, parking efficiency, and control smoothness, encouraging the EV to reach the target pose safely and efficiently~\cite{pbrlmp6,Park02}. 

Racing is a performance-oriented domain on closed circuits with intense vehicle competition. Similar to lane changing and overtaking, racing requires the EV to continuously surpass SVs to improve its position~\cite{Racing01}, but these maneuvers are not constrained by lane boundaries and only need to remain within the racetrack. Accordingly, rRL design emphasizes vehicle dynamics exploitation, and tracking or discovering a time-efficient racing line. 
Most studies directly output steering and acceleration commands, while some further introduce high-level semantic actions to facilitate cooperative or adversarial behaviors~\cite{Racing02, Bb5hrl}. Besides conventional vehicle states and surrounding observations, racing tasks often incorporate track-progress information and reference racing-line guidance~\cite{TraD-RL}, and vehicle-limit-related states such as tire temperature and engine speed~\cite{Bb5hrl}, to facilitate high-speed trajectory optimization and support vehicle control near handling limits. Rewards are usually dominated by progress and lap-time objectives, while overtaking rewards~\cite{Racing01}, collision penalties, and sportsmanship constraints~\cite{I10_Nat_Racing} are introduced to ensure competitive yet reasonable driving behavior. Furthermore, human demonstrations~\cite{Bb5hrl}, curriculum learning~\cite{Racing02}, and MARL frameworks~\cite{LC12_1} have been employed to improve overtaking performance and interaction strategies. Notably, RL agents have achieved human-champion-level performance in racing simulators~\cite{I10_Nat_Racing}.

Off-road driving shifts the focus from traffic interactions to environmental uncertainty and terrain adaptability. Unlike urban scenarios, off-road environments often lack explicit lane boundaries, traffic signs, and structured road geometry, requiring the agent to reason about terrain traversability, irregular obstacles, and vehicle-terrain interactions. Accordingly, observation representations frequently incorporate terrain maps, elevation information, vehicle attitudes, obstacle features, and target-related information~\cite{OffRoad03}. Many studies formulate off-road driving as a path-planning problem, where RL is used to jointly optimize traversability, energy consumption, and travel distance~\cite{OffRoad01,OffRoad02}. Furthermore, model-based RL has been employed to explicitly account for model uncertainty and complex vehicle dynamics. Zhu et al.~\cite{Offroad_traversability} develop a traversability mapping method that integrates terrain geometric and semantic information for RL agent observation augment. Wang et al.~\cite{OffRoad04} trains a probabilistic dynamic model to consider model-uncertainty, thereby improving robustness and adaptability across diverse off-road environments.

\subsection{Integrated Urban Navigation and End-to-End Driving}

Urban navigation combines multiple common driving tasks within a unified environment. Unlike single-task scenarios, the agent must continuously handle car following, lane changing, merging, intersections, roundabouts, pedestrian interactions, and traffic-rule constraints. Consequently, RL-based MoP shifts from learning task-specific policies to developing a unified policy capable of understanding heterogeneous traffic scenarios and coordinating multiple driving objectives. This unified policy must further generalize across diverse and long-tailed urban situations, many of which are difficult to exhaustively cover during training.

To improve generalization across diverse urban scenarios, recent studies have explored various representation learning and training strategies. Curriculum learning gradually increases navigation complexity, from simple route following to dynamic urban environments with randomized initial states and pedestrian interactions~\cite{UN02}. Transformer-based architectures aggregate unordered traffic participants into compact scene representations, imporving policy robustness under varying traffic densities~\cite{UN03}. Latent representation learning, such as VAE-GAN frameworks~\cite{UN04}, compress high-dimensional visual observations and enhences robustness under adverse weather conditions. 

More recently, integrated E2E architectures have emerged to jointly optimize perception, prediction, and planning within a unified policy network, reducing information loss introduced by modular pipelines and improving overall driving performance~\cite{UN05}. E2E RL policies that further directly optimize closed-loop driving behaviors. Anzalone et al.~\cite{UN02} and Morgado et al.~\cite{Morgado2025} further develop a PPO-based E2E RL policy in CARLA simulator, where the agent directly learns continuous driving actions for go-to-point navigation, lane keeping, and basic urban driving. These studies demonstrate the feasibility of using RL to learn integrated perception-control policies in urban simulation environments. Other studies incorporate constraint-related information or auxiliary critics into E2E RL policy learning to improve policy behavior in crowded urban scenarios~\cite{10679712}. More recently, scalable RL planning has been investigated through simpler reward formulations centered on route completion, allowing PPO-based policies to scale to large-sample training across CARLA and nuPlan benchmarks~\cite{pmlr-v305-jaeger25a}. Meanwhile, E2E RL is also moving toward stronger closed-loop training environments and model-assisted learning. Raw2Drive~\cite{yang2026raw2drive} introduces aligned world models to guide raw-sensor policy training and bridge the gap between privileged planning and raw-sensor E2E driving, while RAD~\cite{gao2026rad} constructs photorealistic 3D Gaussian Splatting (3DGS)-based digital environments for large-scale closed-loop traning, improving policy robustness to out-of-distribution (OOD) scenarios. CarPlanner~\cite{CarPlanner} firstly demonstrates that the RL-based planner can surpass both IL- and rule-based state-of-the-arts on the challenging large-scale real-world dataset nuPlan. These studies indicate that E2E RL is evolving from direct sensor-to-control policy learning toward scalable closed-loop optimization.

\section{Lessons for RL-based MoP design}

Most studies on RL-based MoP focus on specific driving tasks. Each driving task typically involves distinct scenario characteristics and task requirements, which significantly affect the RL agent design. The effective application of RL to a particular driving task requires careful consideration of several critical design elements, including the design of the observation input, the action output, the reward function, and the training environment. According to the review in Sec. III, these design components vary substantially across different driving tasks and algorithms. In addition, these manual designs strongly influence subsequent self-learning and policy iterations. This section summarizes and analyzes the patterns of RL model design, aiming to extract lessons learned from various driving tasks and to provide clear guidelines for the application of RL-basedr MoP techniques for AD.

\subsection{Observation Input}

Observation input determines what information is available for policy learning and how the RL agent understand the driving task. An overly large or weakly structured observation space can make RL policy learning inefficient or unstable. Therefore, observation design should not simply pursue the inclusion of more information, but should balance the task sufficiency and representation compactness.

\textbf{\textit{1) Task-driven State Abstraction}}

The required observation abstraction depends strongly on task complexity. For simple tasks such as CF and LK, compact physical features are usually sufficient because the interaction structure and task objective are relatively clear. These features typically include the motion state of the EV, such as position, heading, speed, and chassis states~\cite{stateenhanced,CTDE4,Bb5hrl}, as well as the relative position, speed, and distance of surrounding traffic participants~\cite{LC7_4,RM5,hc4yuankangevolutionary}. Such low-dimensional states simplify the learning problem and accelerate convergence.

As the driving task shifts from ego-centered motion control to interaction-aware decision making, physical features alone may not fully capture the environment information. LC, merging, and intersections, require the agent to understand SV interactions, priority relations, and potential future conflicts. Therefore, observation design should evolve from compact ego/object states toward interaction-aware representations that encode both vehicle-level dynamics and task-relevant context~\cite{stateenhanced,LC7_5,vbrlmp8}.

\textbf{\textit{2) Multimodal and Context-aware Observation}}

Sensor inputs, such as camera images~\cite{vbrlmp4, Park02}, bird's-eye-view (BEV) representations~\cite{LC22}, and LiDAR point clouds~\cite{pbrlmp5,Racing01}, are often introduced to reduce information loss caused by manually selected features. They are particularly useful for complex scenarios where spatial occupancy, occlusion, and heterogeneous traffic participants are difficult to describe using only low-dimensional physical states. In urban navigation and E2E driving, multimodal observations are commonly used to combine abstract physical features with multi-source perceptual and contextual information, including road maps, global routes, traffic lights, stop lines, speed limits, and other traffic-rule information~\cite{UN02,stateenhanced,RD04}. In collaborative tasks such as formation driving and cooperative merging, V2X information can further extend the observation space by providing interaction-related information beyond the ego vehicle's onboard perception range~\cite{RM10_1,InMulti2}.

Nevertheless, multimodal input does not automatically lead to better policy learning. Directly feeding high-dimensional and multi-source observations into RL networks may increase computational complexity and make latent feature extraction more difficult. In many cases, abstract physical features remain necessary even when raw sensor inputs are used, because they provide stable and task-relevant information that is easier for the agent to exploit during trial-and-error learning.

\textbf{\textit{3) Auxiliary Representation}}

Auxiliary representations further convert task-related priors into compact and structured inputs, helping the RL agent understand environment structure, task objectives, and operational constraints. For example, grid maps can encode spatial occupancy~\cite{gridinput}, while risk-related metrics such as TTC or other indicators provide compact safety-relevant abstractions~\cite{vbrlmp8,Racing01}. Risk potential fields can further provide driving-cost information~\cite{OI01}. Historical trajectories can also be used to capture temporal interaction patterns that are not fully reflected by instantaneous states~\cite{LC7_5}. For off-road driving, terrain maps and elevation information become more important because the agent must reason about traversability rather than lane-level traffic structure~\cite{Offroad_traversability}.

Auxiliary representations should be introduced according to the missing information of the task, rather than added as redundant inputs. Effective observation design should convert task-relevant information into compact and robust representations. In this sense, the observation input is not merely a data interface, but a task-driven abstraction layer helping RL agent efficiently learn the relationship among environment state, action consequence, and long-term driving objective.

\subsection{Action Output}{\label{action}}
Action output defines the intervention level of RL in the MoP pipeline. Action space design is determined only by whether an RL algorithm supports discrete or continuous pattern, but also aligned with the planning horizon and motion-control requirements of the driving maneuvers. In general, short-horizon control maneuver favor low-level continuous actions, while interaction-intensive and long-horizon tasks tend to require more abstract, structured, or hierarchical action representations. This makes action output a key interface between policy learning and executable vehicle motion.

\textbf{\textit{1) Action Abstraction}}

A key design principle is that action abstraction should be consistent with the planning horizon and objective of the driving task. For tasks requiring behavior-level reasoning over a relatively long horizon, the RL agent often outputs high-level semantic actions to represent driving intentions. Typical examples include lane-level behaviors such as LK, LCL, and LCR, acceleration/deceleration choices associated with lane changing, overtaking, and merging~\cite{LC7_4,vbrlmp4,hc4yuankangevolutionary}, and priority-related behaviors such as pass, yield, wait, take up, and give up in conflict-zone scenarios~\cite{giveaway,In2_2}. These actions are easy to interpret and can be naturally connected with downstream planning or control modules. However, their coarse granularity may limit maneuver flexibility in highly dynamic scenarios. Restricting the agent to a finite set of discrete actions may prevent the policy from generating near-optimal motion in continuously changing traffic and vehicle-dynamics conditions. Moreover, abrupt transitions among discrete action may introduce jerky driving maneuvers.

When the task requires more flexible short-term vehicle motion control, the action space usually needs to move from discrete semantic commands to continuous commands.  Therefore, many RL-based MoP studies allow the agent to directly output continuous commands, such as steering angle, acceleration, target speed, or pedal-level control~\cite{pbrlmp4,pbrlmp5,pbrlmp6,Racing01}. Such actions provide finer control resolution and greater maneuver flexibility. However, continuous action spaces also increase the difficulty of policy learning and stable execution.

Between semantic decisions and direct control commands, trajectory-level actions provide an intermediate abstraction. The RL agent can generate candidate trajectories by outputting target points, polynomial coefficients, planning distances, or objective-function weights~\cite{LC22,TPA, hc4yuankangevolutionary, hc5yuanevolutionaryits}, which are then converted into executable motion by downstream planning and control modules. When trajectory-level actions either contain discrete-continuous hybrid components themselves or need to be integrated with upper-level discrete semantic behaviors, parameterized-action RL frameworks allow heterogeneous trajectory or planning parameters to be generated synchronously, as in P-DQN~\cite{TraA01}, RL-TPA~\cite{JinNNLS,JinRAL}, and related approaches. These intermediate actions enrich the expressiveness of RL-based MoP while preserving more structure than direct low-level control.

For complex tasks involving multiple action levels, hierarchical or hybrid action spaces can further combine semantic decisions, continuous control actions, and trajectory-related commands. Such structures may be serial, with high-level discrete behaviors followed by low-level control commands, parallel, with heterogeneous commands generated simultaneously, or hybrid combinations~\cite{HA01,vbrlmp8,hl2}. Thus, they serve as structured action abstractions that align long-horizon driving intentions with short-horizon motion execution.

\textbf{\textit{2) Control Granularity}} 

Action's control granularity determines how precisely the RL agent can influence vehicle motion. Coarse-grained actions, such as discrete behavior decisions, reduce the learning difficulty, but their influence on vehicle motion is usually indirect and depends on downstream planning or control modules to generate executable trajectories. Therefore, they may not provide sufficient flexibility for continuously changing traffic interactions and vehicle-dynamic conditions. In contrast, fine-grained continuous actions, such as steering angle and acceleration commands, provide the agent with direct control authority and higher motion-control resolution. However, this tighter coupling between policy output and vehicle motion also enlarges the action space and makes policy learning more sensitive to observation noise and reward design, especially when small observation perturbations lead to large fluctuations in steering or acceleration outputs.

Trajectory-level actions describe a future motion segment rather than an instantaneous actuator-level command. By outputting waypoints, trajectory parameters, or planning-related variables, the agent can characterize intended vehicle motion, which improves interpretability and facilitates safety, feasibility, and smoothness evaluation before execution. However, RL learns from accumulated rewards, making high-dimensional waypoint sequences or numerous trajectory coefficients difficult to optimize. Thus, overly detailed trajectory-level actions may enlarge the action space, thereby slowing convergence, and destabilizing policy learning.

Therefore, action granularity should not be designed by simply pursuing finer control or more detailed trajectory outputs. The desired representation should be sufficiently expressive to characterize the future motion required by the task, but compact enough to support efficient exploration and stable learning. In this sense, effective action granularity design requires balancing motion expressiveness, learning efficiency, interpretability, and downstream feasibility, rather than choosing the finest possible action resolution.

\subsection{Reward Function}

Reward design can significantly influence the performance of the RL agent as it directly informs the loss function required for network updating. Driving is a multi-attribute problem, and these attributes may include the time to reach destination, travel distance, collision, legal compliance, energy consumption, passenger experience, impacts on the traffic environment, etc.~\cite{RD02}. Defining a driving performance metric for an autonomous vehicle involves identifying various attributes and quantitatively describing them, and then combining them into a utility function. Current RL models for AD typically formulate the reward function as a weighted linear combination~\cite{I7_RL_IL}.

\textbf{\textit{1) Reward Attributes}}

Reward attributes define the driving outcomes that the RL policy is expected to optimize. A primary requirement is task completion, because the agent should, as far as possible, accomplish the desired driving tasks~\cite{pbrlmp5,LC7_4,Park02}. On this basis, RL-based MoP further introduces several reward attributes, such as safety, efficiency, comfort, to evaluate the quality of task completion. Safety determines whether the task is completed without collisions or dangerous proximity, and is usually represented by collision or out-of-road penalties~\cite{vbrlmp9,vbrlmp4,pbrlmp6,RL_GT}, as well as surrogate risk metrics such as TTC, THW, relative distance, or risk values~\cite{CF1,RM10_1}. Efficiency reflects whether the task is completed without unnecessary delay or excessive conservatism, and is commonly related to speed, travel time, path length, or other task-progress cost~\cite{vbrlmp9,Racing02,LK3}. Comfort measures whether the generated motion is acceptable for passengers and vehicle execution, typically through acceleration, jerk, lateral acceleration, steering variation, or frequent lane changes~\cite{vbrlmp9, pbrlmp6, stateenhanced, LC22, Park02, hl2}. Traffic compliance evaluates whether the task is completed under valid road-use constraints, including traffic signals, lane rules, speed limits, stop lines, and other scenario-specific regulations~\cite{fearsrl,In4_3,I6lisrl,RD10_r,rightway}.

Reward attributes should not be treated as a fixed checklist, because their relative importance depends on the dominant objective of each task. For example, CF and LK emphasize tracking performance, safety, and smoothness; LC, merging, and intersections require stronger safety and interaction-related rewards; parking focuses on terminal position and heading accuracy; racing prioritizes progress and lap time while penalizing collisions and unreasonable behaviors; and urban navigation requires route completion, traffic-rule compliance, and interaction safety. Therefore, reward design should first clarify what constitutes successful task completion, and then select only the necessary attributes for evaluating behavior quality, since excessive reward terms may increase tuning difficulty, introduce conflicting learning signals, and obscure the primary task objective.

\textbf{\textit{2) Reward Shaping}} 

When reward signals from objective attributes are sparse, it is a natural idea to encourage and indicate seemingly desirable maneuvers in reward functions~\cite{RD02}. For example, encouraging staying near the lane centerline can help a vehicle to quickly learn how to keep on track. However, combining this partially shaped rewards with existing safety rewards may lead the RL agent unexpectedly fall into a local optimum, such as persistently following SVs at a low speed, which is not actually the desired driving behavior. Common shaped rewards via one or more attributes include suggesting zero steering angle~\cite{RD04}, increasing the separation distance with SVs~\cite{I6lisrl}, overtaking other vehicles~\cite{LC12_1}, etc. While reward shaping improves the learning efficiency, it may reduce the achievable performance by subjectively changing the preference order of the reward function. As Russell and Norvig assert~\cite{russell}, \textit{“It is better to design performance metrics according to what one actually wants to be achieved in the environment, rather than according to how one thinks the agent should behave”}. The survey~\cite{RD02} boils it down to a pithy description: \textit{“specify how to measure outcomes, not how to achieve them.”} Despite its theoretical drawbacks, reward shaping remains effective in RL-based MoP methods as of this time. Until a better learning way emerge, reward shaping techniques can enhance driving performance to a certain extent and their potential side effects should be carefully examined. Effective reward design remains an open problem for RL-based MoP and other control tasks.

\textbf{\textit{3) Reward Utilization}}

Most RL-related studies combine multiple reward attributes through weighted summation. Although weight-factor limits can be analytically discussed for simple attributes such as crash, idle, and success~\cite{RD02}, manually designed weights are difficult to generalize across tasks and cannot fully resolve conflicts among safety, efficiency, comfort, and compliance. Therefore, parameter-tuning methods such as GLIS~\cite{GLIS} and inverse reinforcement learning (IRL) methods~\cite{RD06,RD05} have been introduced to optimize reward weights or learn reward values from expert experience.

However, optimizing weights alone often provides limited performance gains, because agents may still favor attributes with larger numerical rewards. More structured approaches therefore separate or adapt reward utilization: Multi-Critic methods use multiple critics or Q-networks to estimate different reward components and balance learning objectives~\cite{JinNNLS, RD07}, while reward machines condition reward transitions on task context to adapt rewards across scenarios~\cite{RD09}. Recent work further shows that simplifying reward utilization can also be effective. Instead of optimizing many additive shaped rewards, CaRL~\cite{CaRL} primarily optimizes route completion and handles infractions through episode termination or multiplicative penalties, improving the scalability of PPO-based driving policy learning.

Recent studies further move beyond manually specified reward terms by using Large-Laguage-Models (LLMs), or learned reward models to assist reward construction and policy optimization. AutoReward~\cite{AutoReward} uses LLMs to generate and iteratively refine reward functions according to task descriptions, environment information, and RL training feedback. Gen-Drive~\cite{GenDrive} trains a scene evaluator as a reward model from vision-language-model (VLM)-assisted pairwise preference data and uses it for trajectory evaluation and RL fine-tuning of diffusion-based driving policies. DriveReward~\cite{DriveReward} builds a generative VLM reward model from a reward-oriented trajectory-evaluation dataset, where front-view images, ego states, navigation commands, and candidate trajectories are annotated with multi-dimensional reward scores. DriveMind~\cite{wasif2025drivemind} uses a dual-VLM reward framework with semantic anchoring and dynamic prompt generation to provide adaptive reward signals. During RL fine-tuning, its predicted PDMS is directly used as the reward signal. IRL-VLA~\cite{IRLVLA} further constructs a lightweight reward world model through IRL and uses it to provide closed-loop reward feedback. These methods shift reward utilization from static hand-crafted weighting toward adaptive reward optimization.

\section{Exploratory efforts to address contemporary challenges}

Although there have been many significant achievements in RL-based MoP, there are still many challenges in applying it to real-world AD systems. Owing to page limitations, we focus on three attributes that have the greatest impact on RL-based MoP for AD, i.e., safety performance, sample efficiency, and generalization capability. Other attributes, such as interpretability and ethics, are not discussed in this survey, and interested readers are referred to \cite{I23gaozhenhai,TRC2024recent}, which address such topics. This section reviews recent exploratory efforts for these three frontier issues and proposes directions for future research. Since promoting sample efficiency and generalization capability share some common technical aspects, we distinguish them according to the  primary motivation for using these techniques to enhance the performance of RL-based MoP.

\subsection{Safety Performance}{\label{safe}}
Safety is a fundamental requirement for AD. However, the RL agent may sometimes prioritize maximizing the overall reward over ensuring safety, especially under conditions where multiple objectives are considered. This can lead to unsafe or even disastrous behaviors, which is the most important hindrance to the application of RL to real-world AD~\cite{CTDE1}. Consequently, an increasing number of researchers have focused on the safety of RL-based MoP methods and have begun to explore the application of Safe RL.

Safe RL is often modeled as the Constraint MDP (CMDP)~\cite{shenboCMDP}, which additionally minimizes safety-related cost $\mathcal{C}_\pi(s)\!=\!\mathbb{E}[ {\textstyle\sum_{t=h}^{H+h}}\gamma^{t-h}c_{t+1}|s_h\!=\!s ]$ while maximizing cumulative reward expectations, where $c_t$ is the safety-related cost value at timestep $t$. The objective of CMDP is to find a policy $\pi_{\theta} \in \Pi_{\mathcal{C}}$ to maximize the expected reward, where $\Pi_{\mathcal{C}}=\{\pi_{\theta} | \mathcal{C}_{\pi}(s) \leq \mathcal{C}_{thres}\}$ represents the safe policy set with a cost threshold $\mathcal{C}_{thres}$. Safe RL applied in the MoP can usually be categorized as: i) Policy objective optimization: This method uses the cumulative cost values on the trajectories to search for safe policies, gradually converging to safe set. ii) Hard safety constraint: Stricter requirements on the safety of each step are imposed during training or testing through predefined constraints. This type of approach can further enhance safety, but is more conservative.

\textit{1) Policy Objective Optimization}

Constrained Policy Optimization (CPO) is frequently used to guide the generation of a safer policy~\cite{cpo3}. Wen et al.~\cite{CPO2} employed parallel CPO agents to collect sufficient safe and feasible experiences for policy updates, reducing unsafe updates in hazardous situations. Lagrangian-based methods transform constrained safety optimization into unconstrained objectives with adaptive multipliers. For example,~\cite{stateenhanced} uses a Lagrangian network to adjust penalties for constraint violations and a feasible value network to evaluate policy feasibility. Lv et al.~\cite{fearsrl} introduce a fear model inspired by the amygdala mechanism to recognize potential risk and optimize policy under fear constraints.

Control-theoretic constraints are also integrated into policy objective optimization.
In~\cite{CLFSRL}, a distance-based Control Lyapunov Functions (CLF) is established, treating the collision probability as a risk factor in the critic with the policy gradient to improve safety. Udatha et al.~\cite{CBFSRL1} implement a probabilistic Control Barrier Function (CBF) and convert it into linear control constraints to ensure safety during policy updates. To reduce dependence on manually specified safety functions, Yang et al.~\cite{CBFSRL2} further learn a barrier function from collected unsafe and initial states.

Uncertainty-aware methods further guide safer exploration and policy updates. In~\cite{nageshrao2022robust}, the RL policy is updated only when its performance confidence exceeds the baseline. Zhang et al.~\cite{uncsrl1} use variance from ensemble critic networks to encourage exploration and to determine when to switch to a rule-based fallback. In~\cite{uncsrl2}, CVaR-based distributional critics support safe  policy updates, with the policy space adaptively expanding when boundary actions are identified as safe.

\textit{2) Hard Safety Constraint}

Setting driving rules or rule-based MoP as a safety filter is an intuitive way to enhance the policy's safety.
Gu et al.~\cite{hc1} combines traditional MoP method with RL, where the safety buffers around obstacles constrain the RL output to collision-free path points. Wang et al.~\cite{hc2} design a CBF-based safety filter to verify RL actions under longitudinal and lateral constraints and predefined traffic rules before execution. Some filters ely on conditional safety checks. Reference~\cite{hc3} uses Linear Temporal Logic (LTL) based on prior safety rules to to evaluate policy safety and trigger a rule-based emergency response when the RL action is unsafe. References~\cite{hc4yuankangevolutionary,hc5yuanevolutionaryits} employ MPC-based longitudinal pre-planning to check whether a safe acceleration can be generated. Otherwise, the unsafe action is masked and fed back to policy update.

Uncertainty can also be used in constraint design for safe RL~\cite{Bb6Exploration}. Aleatoric uncertainty is often reflected by return distributions and can support risk-sensitive trade-offs between safety and efficiency after convergence~\cite{ensemble}. Epistemic uncertainty usually arises from the insufficient training data or OOD scenarios, and is commonly estimated by ensemble value variance. In~\cite{hc7tangxiaolin}, the RL policy reverts to a rule-based policy if the uncertainty evaluated exceeds a safety threshold.  

Prediction information provides a direct way to evaluate long-horizon safety risks. In~\cite{I6lisrl}, actions are mapped to trajectories, whose risks are assessed based on the EV planning trajectory and the SVs' predicted trajectories. High-risk actions are then discarded and replaced by safer alternatives, reducing risky behaviors in real-world lane-change experiments. Moreover, Krasowski et al.~\cite{hcpre2} and Gu et al.~\cite{stateenhanced} further construct safe action sets within prediction-embedded frameworks to replace unsafe RL actions when the learned following strategy fails.
Additionally,~\cite{linansrl1} and~\cite{linansrl2} construct optimization-based filters to guarantee that the agent remains safe at all times, while minimizing modifications to the RL policy. A related work~\cite{safefilter} uses MPC as a filter to to keep the agent within a safe invariant set.

\subsection{Sample Efficiency}
Due to the trial-and-error learning paradigm, RL usually requires a substantial number of interactions to collect reward-feedback experiences and then learn feasible policies. This sample-efficiency problem is particularly evident in the AD MoP field because of the open interaction environment with large state space and hard-to-collect long-tail data, which results in slow driving policy convergence and lower-than-expected driving performance. Several factors further increase the sample demand. i) the complexity of traffic interactions makes it difficult to obtain objective and informative reward signals. ii) many driving tasks exhibit temporal correlations, which can further amplify the effects of delayed rewards. iii) the RL agent must spend a considerable amount of time on constant trial-and-error in the massive exploration space. iv) valuable samples become increasingly difficult to obtain in later training stages. To address these challenges, recent advances have explored how RL agents can acquire more useful driving experience from limited samples, aiming to improve MoP performance and promote the practical deployment of RL-based methods in AD.

\textit{1) Learning from Demonstration (LfD)}

LfD improves sample efficiency by providing prior driving experience before or during RL exploration, especially when reward signals are sparse or the exploration space is too large for random trial and error. Demonstration can be obtained from prior rule expert, human guidance data, or pre-trained policies.

\textit{a) Learning from a rule-based expert:} The RL agent can be simply and directly guided through rule-based model demonstration. Alighanbari et al.~\cite{lfdrule1} use an NMPC controller to generate switchable policy demonstrations, guiding the DDPG to accelerate convergence. In~\cite{lfdrule3}, an expert system composed of constrained iterative LQR and PID controllers is incorporated into RL training. Similarly, Li et al.~\cite{I6lisrl} design a formalized rule-based correction mechanism considering predicted risks, where multi-memory batches are set to store exper-guided experiences for more efficient learning.

\textit{b) Learning from Human-Guidance:} Human guidance can alleviate sample inefficiency by providing human demonstrations or interventions during RL training. Liu et al.~\cite{lfdhg2} combine reward maximization with expert imitation and adaptively sample from both self-exploration and human demonstrations. In~\cite{hc5yuanevolutionaryits}, human online interventions are triggered for unfavorable actions, limiting unsafe exploration and provide demonstrations in complex scenarios. Similarly, Wu et al.~\cite{lfdhg6lvchen24} establish an integrated framework including human/RL action switch mechanism, advantage-based prioritized experience, and human-intervention reward shaping to improve the use of human guidance. Recent studies further propose a human-guided distributional SAC framework that encode intervention information into proxy values for reward-free and sample-efficient real-world policy learning~\cite{Human-in-the-Loop2025}, or integrate real-time takeover demonstrations into a continual RL framework to personalize the deployed driving policy while preserving previously learned RL driving abilities~\cite{Human-Guided2025}.

\textit{c) Learning with a pre-trained policy:} A near-optimal policy can effectively initialize online fine-tuning. Huang et al.~\cite{lfdpt2lvchenBCRL} distill human prior knowledge into an imitative expert model, and then add a KL-divergence penalty to guide online RL toward the expert policy. Shi et al.~\cite{OI01} DAgger to train an IL agent for RL initialization, reducing data demand under sparse rewards. Inspired by the recent success of Group Relative Policy Optimization (GRPO)~\cite{shao2024deepseekmath}, group-wise trajectory sampling has become a practical way to fine-tune pre-trained policies with reward feedback. DiffusionDriveV2~\cite{DiffusionDriveV2} and PlannerRFT~\cite{PlannerRFT} both refine pre-trained diffusion planners using GRPO to improve trajectory quality and exploration efficiency. WorldRFT~\cite{WorldRFT} proposes a planning-oriented latent world model framework and uses GRPO with trajectory Gaussianization and collision-aware rewards to improve safety-critical planning. ReCogDrive~\cite{ReCogDrive} integrates an autoregressive model with a diffusion planner, and then applies Diffusion GRPO to improve safety and comfort. PaIR-Drive~\cite{PaIRDrive} decouples IL and RL into parallel branches, where RL searches trajectories beyond IL priors. R2SE~\cite{R2SE} starts from an IL-pre-trained E2E generalist and applies hard-case-oriented residual reinforced fine-tuning with adapter expansion to improve difficult scenarios while preserving general driving abilities.

\textit{2) Task Differentiation}

It challenging to directly learn an effective driving policy in a complex MoP task. Decomposing the task into different parts is a feasible way. Instead of learning to deal with the whole task directly starting from a complex environment, the agent learns the different sub-tasks in stages. Easier initial tasks can guide the agent toward the final task, thereby reducing learning complexity and improving convergence and sample efficiency~\cite{td2CLSV-TPAMI}.

\textit{a) Curriculum Learning (CL):} CL trains RL agents through tasks with gradually increasing difficulty, helping avoid premature failure in highly complex scenarios. Traditional CL relies on manually designed stages. For example, Shi et al.~\cite{cl1} design three-stage curriculum RL including adaptive cruise control, lane changing, and overtaking tasks with different reward functions. Anzalone et al.~\cite{UN02} progress from static to complex traffic and weather conditions through five stages to learn complex behaviors. Research on automatic curriculum generation has emerged to overcome the limitations of manual curriculum design. Banerjee et al.~\cite{Racing02} used Bayesian optimization to automatically select curriculum through probabilistic inference on curriculum-reward functions. Niu et al.~\cite{cl6} dynamically estimate failure probabilities and resample historical scenarios to provide real-time curriculum adaptation. Reference~\cite{cl7} further uses LTL progression to provide task-progress information for RL exploration, although its application to complex AD MoP tasks remains limited.

\textit{b) Transfer Learning (TL):} TL leverages knowledge reuse techniques to utilize knowledge learned from related tasks. Originally, TL was intended to effectively transfer policies to new environments for better generalizability, which will be described in later Sec~\ref{SecVC}. Meanwhile, TL also contributes to sample efficiency, by initializing the policy using prior knowledge from related tasks.
Shu et al. \cite{tl3} improve the control performance and learning efficiency of the Dueling DQN through three transfer rules. Zhou et al.~\cite{zhou2025knowledge} train a teacher policy in a simpler environment and guide student learning in more complex highway lane-change scenarios. Huang et al.~\cite{SafeTransferRL} train a SAC model in a simple highway environment to support target-domain learning through adaptive intervention, reward shaping, and policy-ratio reweighting. TL reduces target-task exploration costs, but poorly aligned source knowledge may cause negative transfer.

\textit{c) Hierarchical Learning:} Hierarchical learning architecture (as discussed in Sec~\ref{action}) decomposes complex MoP tasks at the action-output level, thereby reducing the difficulty of direct policy learning~\cite{cl7}. For instance, directly learning steering command alone may cause large deviations from the lane centerline, because RL agents may not quickly distinguish lane-changing and lane-following behaviors~\cite{hl1}. By combining high-level discrete semantic behaviors with low-level control commands, hierarchical methods can provide clearer driving objectives and more precise motion control\cite{hl2}. Xia et al. \cite{hl4pa} further couple high-level behaviors with low-level continuous actions in a parameterized action space, where fine-grained controls are computed from coarse-grained decisions.

\textit{3) Promoting Exploration}

Efficiently exploring the environment and gathering informative experiences is also important for accelerating learning toward the optimal policy~\cite{Bb6Exploration}.
Uncertainty-oriented exploration generally considers epistemic uncertainty and aleatoric uncertainty, similar to the safety-related discussed in Sec~\ref{safe}. InDRiVE~\cite{InDRiVE} uses latent ensemble disagreement as a proxy for epistemic uncertainty to drive the agent toward under-explored driving situations. Both types of uncertainty are considered in~\cite{pe4} to enhance the robustness of exploration against environmental noise.
Intrinsic motivation-oriented exploration heuristically utilizes reward-agnostic information to promote exploration. In the absence of an explicit feedback, the RL agent can use intrinsic motivation to evaluate the quality of its actions. Ma et al. \cite{pe6Curiosityhuang} introduce curiosity-driven exploration, where curiosity is measured by the prediction error of forward dynamics. Wu et al.~\cite{haoqiracing} also generate an intrinsic reward to encounter the RL agent to explore environment, improving exploration efficiency. 

\subsection{Generalization Capability}{\label{SecVC}} As noted in Sec \ref{review}, most studies have been conducted in low-cost simulation environments tailored to single-task settings. However, task variability can lead to policy failure when applied across different environments. Furthermore, owing to the inherent incomplete limitations of the RL training process, it has poor generalization ability in rare scenarios \cite{gen0}. The ability for long-term multi-task learning is required to enhance the generalization ability of RL agents to the variety of ever-changing complex scenarios that real-world AD applications may face. The generalization ability includes both the policy adaptation cross various driving tasks, and the robustness in response to perturbations during the execution~\cite{robustRL}. Some cutting-edge techniques explored for the generalization capability of RL, but they have not yet been well applied in the field of MoP in AD.

\textit{1) Knowledge Transfer}

\textit{a) Transfer Learning:} Knowledge reuse in TL can improve cross-task and cross-domain generalization, especially for bridging the sim-to-real gap in AD. Kevin et al.~\cite{gentl2} combine domain adaptation and domain randomization techniques by integrating virtual training with real-world data to reduce the sim-to-real transfer gap. Hieu et al.~\cite{Gener03} pre-train on demonstration data using a combination of temporal difference and supervised loss, and then continuously update the policy by incorporating demonstration data with newly collected data, resulting in strong performance across different road conditions and weather conditions. 

\textit{b) Meta Learning:} Meta-RL aims to learn a broadly adaptable policy that can rapidly adjust to new driving tasks or contexts~\cite{genML01}. It usually contains an inner loop for task-specific adaptation and an outer loop for extracting transferable knowledge from multiple tasks. Xing et al.~\cite{genML03} use meta-RL to improve autonomous driving policy adaptation to different drivers. Deng et al.~\cite{genML04} further combine parallel unfolding, multi-task objectives, and a two-stage constraint adaptation strategy to achieve rapid adaptation to new urban driving tasks by reusing meta-training data.

\textit{c) Continual Learning (CRL):} CRL aims to adapt new environments without forgetting previously acquired knowledge~\cite{genContL01}. CRL requires an appropriate balance between the old and new tasks, with adequate generalizability to accommodate their distributional differences. Wei et al.~\cite{genContL03} propose a shared feature extractor with an EWC loss to mitigate catastrophic forgetting and perform velocity control tasks across different environments. Yang et al.~\cite{Human-Guided2025} further develop a human-guided continual learning framework for personalized AD, where real-time human takeover demonstrations update the deployed policy, while priority experience memory-enabled elastic weight consolidation mitigates catastrophic forgetting of fundamental driving abilities.

\textit{2) Policy Stability} 

\textit{a) Disturbance Robustness:} Robust RL focuses on learning policies that remain reliable under external disturbance, such as model mismatch and environmental perturbations. It is often formulated as a two-player zero-sum Markov game, where an adversarial agent trains alongside the ego agent to maximize disturbances. He et al. \cite{dr1} develop an adversarial agent that maximizes the Jensen-Shannon divergence between trained policy and original policy under observation disturbances, and use it as a constraint in Lagrangian policy optimization to improve robustness against observation disturbances. Similarly, the White-Box Adversarial Attack technique is employed to amplify the disturbance of each observation~\cite{RM7_2}.

\textit{b) Uncertainty Adaptation:} Uncertainty can also support RL generalization performance under underrepresented trained scenarios~\cite{ua3}. Lutjens et al.~\cite{ua1} use MC-Dropout and Bootstrapping to achieve parallelized epistemic uncertainty estimation to promote more cautious actions in unknown environments, thereby improving policy generalization. In addition, Hoi et al.~\cite{ua2} propose a risk-conditioned distributional SAC method that learns aleatoric-uncertainty-aware risk-sensitive policies and allows risk-level adjustment without retraining. Hoel et al.~\cite{ensemble} propose Ensemble Quantile Networks (EQN), where ensemble-based epistemic uncertainty supports lower-risk action selection, while quantile functions implicitly capture aleatoric uncertainty to balance risk and efficiency. Additionally, high-uncertainty RL policies can be replaced with more stable baseline policies~\cite{hc7tangxiaolin}, enabling timely adaptation to changes in the environment.

\textit{3) Foundation-Model Enhancement}


Benefiting from large-scale pre-training, foundation models (FMs) possess broad commonsense knowledge, semantic understanding, and emergent reasoning capabilities, thereby providing a new direction for improving the generalization of RL-based MoP~\cite{Dilu2024}. Existing exploratory studies mainly follow two directions: foundation-model-guided RL and Reinforce Fine-tuning (RFT) of FMs.

\textit{a) FM-Guided RL:}
Recent exploratory studies use LLMs and VLMs to guide RL policy learning through semantic reasoning, reward feedback, safety evaluation, and action suggestions~\cite{I31LLMRL}. Some methods attempt to inject LLM priors into the RL training process. For example, \cite{sr8llmrepre} uses LLMs to generate scenario-related state representations and intrinsic rewards, aiming to improve RL adaptability to complex tasks. HCRMP~\cite{LiNeurIPS} leverages LLM generate semantic observation augment and reward preference during training. \cite{Highwayllm} leverages LLM to make safe, collision-free, and explainable predictions for the next states, thereby constructing a trajectory for the ego-vehicle and \cite{pang2024LLMRL} introduce an LLM-based driving expert to provide constraints and intervened interactions guiding the learning process. Found-RL~\cite{FoundRL} further explores asynchronous VLM inference, action guidance, and CLIP-based reward shaping to distill semantic driving knowledge into lightweight RL policies.

World models (WMs) offer another exploratory route for improving RL generalization by enabling close-loop future imagination beyond the collected data. Policy-conditioned imagined rollouts allow agents to evaluate action consequences and train on long-tail scenarios without direct environment interaction. Think2Drive~\cite{li2024think2drive} uses a learned world model as a neural simulator to train the driving planner efficiently. Raw2Drive~\cite{yang2026raw2drive} aligns WMs to transfer privileged rollout knowledge to policy learning. AdaWM~\cite{wang2025adawm} further studies distribution shift in world-model-based RL and adapts the policy or dynamics model through mismatch-aware fine-tuning. VLM-SAFE~\cite{VLMSAFE} couples imagined rollouts with VLM-derived semantic safety scores, enabling safe actor-critic learning under both predicted dynamics and semantic risk guidance. These attempts suggest that world-model-guided RL is moving from imagined-scenario training toward adaptive fine-tuning and safety-aware semantic imagination.

\textit{b) RFT for FMs:}
Another emerging direction is to use RL to optimize foundation driving models themselves. Unlike FM-guided RL, where LLMs or VLMs provide external guidance for an RL policy, this direction directly fine-tunes VLM/VLA-based driving models with reward feedback, aiming to enhance driving-oriented reasoning and planning performance beyond imitation-based supervised fine-tuning (SFT). 

Recent studies first explore reasoning--planning alignment. AutoVLA~\cite{AutoVLA} unifies
reasoning and action generation within a single autoregressive generation model, and then applies GRPO to improve planning performance and reduce unnecessary reasoning. Drive-R1~\cite{DriveR1} aligns visual-grounded reasoning with trajectory planning through rewards on trajectory accuracy and meta-action correctness. OpenREAD~\cite{zhang2025openread} further introduces an LLM-as-critic mechanism, extending RFT from verifiable trajectory outputs to open-ended driving reasoning. MindDriver~\cite{zhang2026minddriver} and OmniDrive-R1~\cite{zhang2026omnidrive} move this direction toward process-level optimization by respectively improving progressive multimodal reasoning and reinforcement-driven visual grounding. Some advances introduce world model magination-guided refinement. For example, ExploreVLA~\cite{sheng2026explorevla} proposes a unified understanding-and-generation framework that leverages world modeling to simultaneously enable meaningful exploration and provide dense supervision via image prediction uncertainty as an intrinsic reward for safe exploration. These methods suggest that RFT can optimize not only final actions, but also the imagined future consequences that support action evaluation. VLA-World~\cite{wang2026learningVLAWM} unifies predictive imagination and reflective reasoning in a VLA world model, where GRPO-based RL optimizes perception, short-term prediction, visual generation, action prediction, and trajectory planning strengthen the advanced reasoning and decision-making capabilities. CausalDrive~\cite{yan2026causaldrive} further develops a controllable real-time causal world renderer that supports reactive counterfactual simulation and large-scale RL post-training through video-based reward generation.
Recent efforts also address failure correction and deployment-oriented adaptation. ELF-VLA~\cite{luo2026unleashing} injects structured failure feedback from a teacher model and reuses corrected samples during GRPO training to overcome the performance plateau in long-tail scenarios.  Overall, RL-based FM fine-tuning is evolving from trajectory-level reward optimization toward reasoning-process alignment, imagination-guided refinement, and failure-aware correction.

\subsection{Open Challenges and Outlook}


\textit{1) Action Representation for Long-Horizon Planning}

Most existing RL-based MoP methods still formulate actions mainly as low-dimensional one-step decisions or short-horizon commands. This design simplifies policy optimization, but it compresses long-horizon driving intentions into local actions and limits the policy’s ability to represent complete, fine-grained trajectories and interaction-aware planning behaviors. Directly learning trajectory-level actions is also difficult because the trajectory dimension, temporal dependency, physical feasibility constraints, and credit-assignment complexity all increase with the planning horizon. Existing hierarchical or parameterized action spaces provide useful intermediate solutions, but their ability to support precise closed-loop trajectory optimization remains limited.

Such bottleneck also limits recent RL fine-tuning of E2E models or driving FMs. Many RFT methods mainly rely on reward-based sampling refinement, or reasoning-format optimization over trajectories generated by pre-trained models, thereby aligning the policy toward higher-reward distribution. However, they do not yet provide a principled mechanism for complete trajectory refinement under dynamic constraints. Existing attempts such as Decision Transformer~\cite{DecisionTransformer}, action tokens~\cite{AutoVLA}, and action chunks~\cite{li2026Qchunk} offer promising inspirations for modeling temporally extended actions. Further exploring these action representations with differentiable optimization may enable RL-based MoP to move from selecting candidate trajectories toward executable long-horizon planning and autonomous closed-loop refinement.

\textit{2) Knowledge-Grounded and Physics-Informed Learning}

Improving sample efficiency remains a fundamental challenge for RL-based MoP, since pure trial-and-error learning is often slow in sparse-reward, long-horizon, and long-tail driving scenarios. Embedding prior knowledge into the learning process can narrow the search space, provide denser learning signals, and guide the agent away from unreasonable actions. Physical knowledge is particularly important because RL-based MoP ultimately aims to generate executable driving actions rather than only interpret driving scenes. The key challenge is how to make the agent internalize such knowledge during closed-loop action generation. 

Even with FM-enhanced RL, this issue remains unresolved. Although VLMs, VLAs, and WMs can provide semantic knowledge, high-level reasoning, or future imagination, their physical understanding of dynamics, action consequences, and closed-loop feasibility remains limited. Moreover, current methods still lack an effective mechanism to transfer such knowledge into downstream RL fine-tuning or action-generation heads. 

Future research may explore physics-informed RL frameworks where physical priors support policy improvement and action generation~\cite{PIRLSurvey}. Promising directions may include physics-aware action-conditioned consequence prediction, dynamics-consistent representations, differentiable rollout optimization. These designs may help generated physically plausible and interpretable actions, even in long-tail or unseen scenarios with limited data coverage.

\textit{3) Risk Evaluation and Safety--Performance Accommodation}

Constructing safety-related objectives or constraints for RL-based MoP first requires a risk representation that can indicate potential danger before the safety boundary is actually violated. However, driving risk is a latent quantity, making accurate risk evaluation difficult. Mechanism-based representations, such as surrogate safety measures or risk fields, are usually limited by handcrafted assumptions. Learning-based methods can extract features from the data to fit this assessment, but collision and near-collision events are rare, and dense frame-level risk labels are generally unavailable. Recent attempts such as GSSM~\cite{GSSM} provide an inspiring direction by indirectly representing risk through deviations from context-dependent naturalistic interaction distributions, yet statistical abnormality is not necessarily aligned with physical danger or collision relevance. Therefore, future research should further develop advanced learning-based risk evaluation methods and investigate how these risk representations can be effectively integrated into safe RL frameworks, supporting proactive constraint construction, risk-sensitive policy optimization, and safer exploration in interactive driving scenarios.

Another challenge is accommodating safety assurance and driving performance. Overly conservative constraints may reduce driving efficiency, while weak penalties may result in collided interaction. Consequently, establishing a quantifiable and dynamically adaptive equilibrium  between safety constraint and motion efficiency for objective adjustment, constraint activation, and action refinement, while assure safety guarantee, remains a critical challenge requiring in-depth investigation.

\textit{4) World Modeling for Exploration and Inference}

WMs provide a promising route for improving both exploration efficiency and generalization in RL-based MoP~\cite{li2024think2drive}. Compared with conventional simulators, data-driven WMs may also better capture realistic visual appearance, traffic interaction patterns, and scenario diversity, by learning the evolution of driving scenes from large-scale data. More importantly, WMs can generate diverse corner cases and long-tail interactive scenarios, allowing the policy to explore safety-critical situations in a low-cost virtual space rather than relying only on rare real-world events or handcrafted simulation.

Beyond data generation for policy training  WMs can also can serve as an imagination module for action-conditioned rollout~\cite{jiang2025world4rl}. Instead of reacting only to the current observation, an RL policy can evaluate candidate actions by rolling out possible future scene evolution and estimating their consequences before execution. This enables more proactive planning under uncertainty and provides a natural interface for risk evaluation, safety constraint activation, and action refinement. 

However, most of current WMs focus on perceptual scene generation or visual future prediction, while much of the physical information most critical for MoP are still lacked~\cite{wang2026WAM}. Long-horizon imagination may further suffer from error accumulation or generative collapse, and high inference cost also limits real-time deployment. Future research may therefore move toward physics-aware world-action models that jointly learn perceptual imagination, scene semantics, physical evolution, and policy actions. Such joint modeling can better couple predictive state modeling with action generation, thereby providing more direct support for RL policy optimization and inference-time decision-making, forming a more closed-loop foundation for long-horizon, risk-aware MoP.

\textit{5) Continuous Policy Evolution and Evaluation}
Long-term deployment of RL-based MoP requires policies to continuously adapt to non-stationary and coupled real-world driving scenarios. Static offline-trained policies are difficult to maintain reliable performance under such non-stationary conditions, while simply fine-tuning policies with newly collected data may cause catastrophic forgetting, safety regression, or unstable behavior updates. Thus, the agent needs to flexibly and efficiently update and evolve to cope with edge cases encountered during driving. Although continual learning and cross-embodiment adaptation have been explored in some robotic research~\cite{abe2026crossEmbodi}, their direct transfer to AD remains limited because driving policies must satisfy strict safety constraints, interact with human road users, and complian traffic-rule. Future research should therefore investigate advance CRL frameworks tailored to MoP while avoiding catastrophic forgetting of general driving skills.

Meanwhile, continuous evolution also calls for new evaluation protocols. Static validation or testing phases may not effectively reflect the performance of continuous evolution. Future benchmarks should therefore assess both plasticity and retention during continuous policy evolution. Designing compact yet comprehensive evaluation criteria for the continual evolution of RL-based MoP remains an important challenge.

\section{Conclusion}

With its ability to explore and optimize policies in complex, dynamic decision-making tasks, reinforcement learning (RL) has emerged as a promising approach for addressing motion planning (MoP) challenges in autonomous driving (AD). This survey provides a comprehensive review of RL-based MoP for AD, focusing on lessons learned from the driving task perspective. We outline the basic theory of RL methodologies, and then delve into their applications in MoP for diverse driving tasks. Scenario-specific features and task requirements are analyzed to illuminate their influence on RL design. On this basis, we summarize key experiences and extract insights for future implementations. Furthermore, we discuss three key frontier issues in RL-based MoP for AD, summarize how some representative emerging technologies are trying to solve them (especially over the past three years), and propose related open issues and future outlooks.

We observe that other approaches and technologies in the field of artificial intelligence are crucial for facilitating the development of RL-based MoP. Future research directions will explore the integration of these advanced methods with frontier issues to promote RL to build AD systems with a better understanding of the world.

\bibliographystyle{ieeetr}
\bibliography{attached}

\vfill

\end{document}